\documentclass{article}

\usepackage[usenames,dvipsnames]{color}
\usepackage{pifont}
\usepackage{graphicx}
\usepackage{subfigure}
\usepackage{natbib}
\usepackage{algorithm}
\usepackage{algpseudocode}
\usepackage{amsfonts}
\usepackage{amssymb,amsmath}
\usepackage{amsthm}
\usepackage{filecontents}

\usepackage{tikz}
\usetikzlibrary{arrows,snakes,calc}
\usetikzlibrary{fit}

\algnewcommand\algorithmicforeach{\textbf{for each}}
\algdef{SE}[FOR]{ForEach}{EndFor}[1]{\algorithmicforeach\ #1\ \algorithmicdo}{\algorithmicend\ \algorithmicfor}

\newcommand{\fitMe}[1]{\resizebox*{\textwidth}{!}{#1}}

\DeclareMathOperator*{\argmax}{\arg\!\max}

\newtheorem*{mytheorem}{Integrality Theorem}
\newtheorem{myproblem}{Problem}

\title{Evaluation Measures for Hierarchical Classification: a unified view and novel approaches}

\author{Aris Kosmopoulos$^{1,2}$, Ioannis Partalas$^3$\\Eric Gaussier$^3$, Georgios Paliouras$^1$ 
       Ion Androutsopoulos$^2$ \vspace{0.2in}
       \\ 
       $^1$National Center for Scientific Research ``Demokritos''\\
       $^2$Athens University of Economics and Business,\\
              Athens, Greece\\
       $^3$Laboratoire d'Informatique de Grenoble,\\
        Univesit\'{e} Joseph Fourier,\\
       Grenoble, France\\
       \{akosmo,paliourg\}@iit.demokritos.gr\\ \{ioannis.partalas,eric.gaussier\}@imag.fr, ion@aueb.gr
}

\begin{document}


\date{}

\maketitle
\begin{abstract}
Hierarchical classification addresses the problem of classifying items into a hierarchy of
classes. An important issue in hierarchical classification is the evaluation
of different classification algorithms, which is complicated by the hierarchical relations among the classes.
Several evaluation measures have been proposed for hierarchical classification
using the hierarchy in different ways. This paper studies the problem of evaluation in 
hierarchical classification by analyzing and abstracting the key components of the
existing performance measures. It also proposes two alternative generic views of hierarchical evaluation
and introduces two corresponding novel measures. The proposed measures, along with the state-of-the-art ones,
are empirically tested on three large datasets from the domain of text classification. The empirical results illustrate
the undesirable behavior of existing approaches and how the proposed methods overcome most of these methods
across a range of cases.
\end{abstract}

\section{Introduction}
Hierarchical classification addresses the problem of classifying items into a hierarchy of classes.
In past years mainstream classification research did not place enough emphasis on the presence of relations between the classes, in our cases hierarchical relations. This is gradually changing and more effort is put into hierarchical classification in particular, partly because
many real-world knowledge systems and services use a hierarchical scheme to organize their data (e.g. Yahoo, Wikipedia).
Research in hierarchical classification has become important, because flat classification algorithms are ill-equipped to address large scale problems
with hundreds of thousands of hierarchically related classes. Promising initial results on large-scale problems show that hierarchical classifiers can be effective in improving information retrieval \citep{kosmopoulos10}.

Many research questions in hierarchical classification remain open. An important
issue is how to properly evaluate hierarchical classification algorithms.
While standard flat classification problems have benefited from established measures such as precision and recall, there are no established evaluation measures for
hierarchical classification tasks, where the assessment of an algorithm becomes more
complicated due to the relations among the classes.
For example, classification errors in the upper levels of the hierarchy (e.g. when wrongly classifying a document
of the class \texttt{music} into the class \texttt{food}) are more severe than those in deeper levels
(e.g. when classifying a document from \texttt{progressive rock} as \texttt{alternative rock}).
Several evaluation measures have been proposed for hierarchical classification (HC) \citep{Costa07,Sokolova09}
using the hierarchy in different ways. Nevertheless, none of them is widely adopted, making it very difficult to compare
the performance of different HC algorithms.

A number of comparative studies of HC performance measures have been published in the literature. An early study can be found in \citep{Sun03}, which is limited to
a particular type of graph-distance measures. A review of HC measures is presented in \citep{Costa07}, focusing on single-label tasks and without providing
any empirical results; in multi-label tasks each object can be assigned to more than 
one classes, e.g. a newspaper article may belong to both \texttt{politics} and \texttt{economics}. In \citep{Nowak10} many multi-label evaluation measures 
are compared, but the role of the hierarchy is not emphasized. Finally, \cite{Brucker11} 
provide a comprehensive empirical analysis of HC performance measures, but they focus on the evaluation of clustering methods  rather than classification ones. While these studies provide  interesting insights, they all miss important aspects of the problem of evaluating HC algorithms. In particular, they do not abstract the problem in order 
to describe existing evaluation measures within a common framework.

The work presented here addresses these issues by analyzing and abstracting the key components of
existing HC performance measures. More specifically:

\begin{enumerate}
\item It groups existing HC evaluation measures under two main types and provides a generic framework for each type, based on flow networks
and set theory.
\item It provides a critical overview of the existing HC performance measures using the proposed
framework.
\item It introduces two new HC evaluation measures that address important deficiencies of state-of-the-art measures.
\item It provides comparative empirical results on large HC datasets from text classification with a variety of HC algorithms.
\end{enumerate}

The remainder of this paper is organized as follows. Section 2 introduces the problem of HC,  presents general requirements for HC measures and
the proposed frameworks. Furthermore, it presents existing HC evaluation measures using the proposed frameworks and introduces two new measures that address problems the state-of-the-art measures have. Section 3 presents a case study comparison and analysis
of the proposed measures and the existing ones. Section 4 describes the empirical setting and data of the empirical analysis of the measures and Section 5 presents and discusses the empirical results. Finally, Section 6 concludes and summarizes remaining open issues.

\section{A Framework of Hierarchical Classification Performance Measures}
This section presents a new framework within which HC performance measures
can be described and characterized. Firstly, supporting notation is defined and then the general
requirements for the evaluation are presented and discussed, based on interesting problems that appear in
hierarchical classification. We then proceed with the presentation of the proposed framework, which is used
in further sections to describe and analyze the measures.

\subsection{Notation}
In classification tasks the training set is typically denoted
as $S=\left\lbrace (\textbf{x}^i,\textbf{y}^i)\right\rbrace_{i=1}^{n}$,
where $\textbf{x}^i$ is the feature vector of instance
$i$ and $\textbf{y}^i \subseteq C$ is the set of classes in which the instance belongs,
where $C=\left\lbrace c_1,\ldots ,c_K \right\rbrace$.

\begin{figure}[ht]
\centering
\subfigure[Tree] {\label{fig:tree} 
\begin{tikzpicture}[-,>=stealth',shorten >=1pt,auto,node distance=3cm, 
  thick,main node/.style={circle,fill=blue!20,draw,font=\sffamily\Large\bfseries, minimum size=14mm, scale=0.4}]

\node[main node] (1) {0};
\node[main node] (2) [below left of=1, xshift=-20mm] {1};
\node[main node] (3) [below of=1, yshift=9mm] {2};
\node[main node] (4) [below right of=1, xshift=20mm] {3};
\node[main node] (5) [below left of=2] {1.1};
\node[main node] (6) [below right of=2] {1.2};
\node[main node] (7) [below of=3, yshift=9mm] {2.1};
\node[main node] (8) [below left of=4] {3.1};
\node[main node] (9) [below of=4, yshift=9mm] {3.2};
\node[main node] (10) [below right of=4] {3.3};
\node[main node] (11) [below left of=9] {3.2.1};
\node[main node] (12) [below right of=9] {3.2.2};

\path[every node/.style={font=\sffamily\small}]
    (1) edge node [left] {} (2)
	edge node [] {} (3)
	edge node [right] {} (4)
    (2) edge node [left] {} (5)
	edge node [right] {} (6)
    (3) edge node [] {} (7)
    (4) edge node [left] {} (8)
	edge node [] {} (9)
	edge node [right] {} (10)
    (9) edge node [left] {} (11)
	edge node [right] {} (12);
\end{tikzpicture}
}\qquad
\subfigure[DAG] {\label{fig:dag}
\begin{tikzpicture}[-,>=stealth',shorten >=1pt,auto,node distance=3cm, 
  thick,main node/.style={circle,fill=blue!20,draw,font=\sffamily\Large\bfseries, minimum size=14mm, scale=0.4}]

\node[main node] (1) {0};
\node[main node] (2) [below left of=1, xshift=-20mm] {1};
\node[main node] (3) [below of=1, yshift=9mm] {2};
\node[main node] (4) [below right of=1, xshift=20mm] {3};
\node[main node] (5) [below left of=2] {1.1};
\node[main node] (6) [below right of=2] {1.2};
\node[main node] (7) [below of=3, yshift=9mm] {2.1};
\node[main node] (8) [below left of=4] {3.1};
\node[main node] (9) [below of=4, yshift=9mm] {3.2};
\node[main node] (10) [below right of=4] {3.3};
\node[main node] (11) [below left of=9] {3.2.1};
\node[main node] (12) [below right of=9] {3.2.2};

\path[every node/.style={font=\sffamily\small}]
    (1) edge node [left] {} (2)
	edge node [] {} (3)
	edge node [right] {} (4)
    (2) edge node [left] {} (5)
	edge node [right] {} (6)
    (3) edge node [] {} (7)
edge node [left] {} (6)   
    (4) edge node [left] {} (8)
	edge node [left] {} (7)
	edge node [] {} (9)
	edge node [right] {} (10)
    (9) edge node [left] {} (11)
	edge node [right] {} (12);
\end{tikzpicture}
}
\caption{A tree and a DAG class hierarchy.}
\label{fig:hier}
\end{figure}
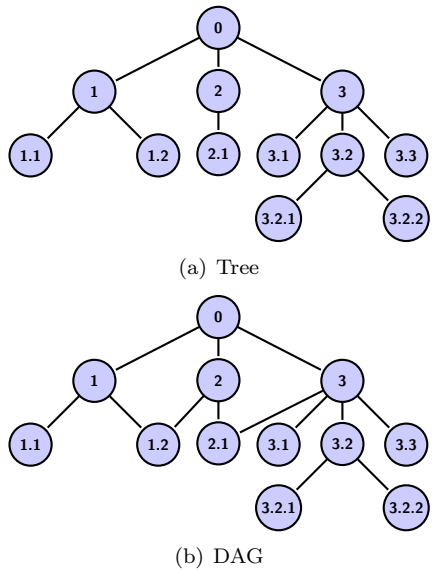

In contrast to flat classification, where the classes are considered unrelated, in HC the classes are organized in taxonomies.
The taxonomies are usually either trees, in which case nodes (classes) have a single parent each, or directed acyclic graphs (DAGs), 
in which case nodes can have multiple parents; see Figures \ref{fig:tree} and \ref{fig:dag} respectively.
In some cases, hierarchies may also be cyclic graphs. In all cases the hierarchy
imposes a parent-child relation among the classes, which
implies that an instance belonging in a specific class, \emph{also belongs 
in all its ancestor classes}. A taxonomy is thus usually defined as a pair
$(C,\prec)$, where $C$ is the set of all classes \citep{Silla11} and
$\prec$ is the \texttt{subclass-of} relationship with the following properties:\footnote{Without loss of generality,
we assume a \texttt{subclass-of} relationship among the classes, but in some cases a different
relationship may hold, for example \texttt{part-of}. We assume however, that the three properties always hold for the relationship.}

\begin{itemize}
\item Asymmetry: if $c_i \prec c_j$ then $c_j \nprec c_i$ for every $c_i$, $c_j$ $\in C$.
\item Anti-reflexivity: $c_i \nprec c_i$ for every $c_i \in C$.
\item Transitivity: if $c_i \prec c_j$ and $c_j \prec c_k$, then $c_i \prec c_k$ for every $c_i$, $c_j$, $c_k$ $\in C$.
\end{itemize}

In graphs with cycles, only the transitivity property holds. In this article we consider only hierarchies without cycles and we denote the descendants and ancestors of a class $c \in C$ as $De(c)$ and $An(c)$, respectively. The parents of a class $c$ are denoted as $Pa(c)$. Finally, we assume that an instance can be classified in any class of the hierarchy, and not only in the leaf classes.

\subsection{General Problems in Hierarchical Classification Evaluation}\label{sec:problems}
The commonly used measures of precision, recall, F-measure,
accuracy etc. are not appropriate for HC, due to the relations that exist among the classes. A hierarchical
performance measure  should use the class hierarchy in order to evaluate properly HC algorithms.
In particular, one must account for several different types of
error according to the hierarchy. For example, consider the tree hierarchy in Figure \ref{fig:tree}.
Assume that the true class for a test instance  is \texttt{3.1} and that two different classification systems output
\texttt{3} and \texttt{1} as the predicted classes. Using flat evaluation measures, both systems
are punished equally, but the error of the second system is more severe as it makes a prediction
in a different and unrelated sub-tree.

\begin{figure}[ht]
\centering
\subfigure[Over-specialization]{
\begin{tikzpicture}[-,>=stealth',shorten >=1pt,auto,node distance=3cm,
  thick,main node/.style={circle,fill=blue!20,draw,font=\sffamily\Large\bfseries, minimum size=11mm, scale=0.7}]

\node[main node] (1) {};
\node[main node] (2) [below of=1,,yshift=9mm] {A};
\node[main node] (3) [below of=2,,yshift=9mm] {B};

\path[every node/.style={font=\sffamily\small}]
    (1) edge node [] {} (2)
    (2) edge node [right] {} (3);

\node[draw=red,inner sep=0pt,thick,circle,fit=(2)] {};
\node[draw=blue,inner sep=0pt,thick,fit=(3)] {};
\end{tikzpicture}\label{fig:ph2}}
\qquad
\subfigure[Under-specialization]{
\begin{tikzpicture}[-,>=stealth',shorten >=1pt,auto,node distance=3cm,
  thick,main node/.style={circle,fill=blue!20,draw,font=\sffamily\Large\bfseries, minimum size=11mm, scale=0.7}]

\node[main node] (1) {};
\node[main node] (2) [below of=1,,yshift=9mm] {A};
\node[main node] (3) [below of=2,,yshift=9mm] {B};

\path[every node/.style={font=\sffamily\small}]
    (1) edge node [] {} (2)
    (2) edge node [right] {} (3);

\node[draw=red,inner sep=0pt,thick,circle,fit=(3)] {};
\node[draw=blue,inner sep=0pt,thick,fit=(2)] {};
\end{tikzpicture}\label{fig:ph3}}\qquad
\subfigure[Alternative paths]{
\begin{tikzpicture}[-,>=stealth',shorten >=1pt,auto,node distance=3cm,
  thick,main node/.style={circle,fill=blue!20,draw,font=\sffamily\Large\bfseries, minimum size=11mm, scale=0.7}]

\node[main node] (1) {A};
\node[main node] (2) [below left of=1] {B};
\node[main node] (3) [below right of=1] {C};
\node[main node] (4) [below of=3,,yshift=9mm] {D};
\node[main node] (5) [below left of=4] {E};

\path[every node/.style={font=\sffamily\small}]
    (1) edge node [left] {} (2)
	edge node [right] {} (3)
    (2) edge node [right] {} (5)
	(3) edge node []{}(4)
	(4) edge node[left]{}(5);
	
	\node[draw=red,inner sep=0pt,thick,circle,fit=(5)] {};
	\node[draw=blue,inner sep=0pt,thick,fit=(1)] {};
\end{tikzpicture}
\label{fig:ph1}}

\subfigure[Pairing problem]{
\begin{tikzpicture}[-,>=stealth',shorten >=1pt,auto,node distance=3cm,
  thick,main node/.style={circle,fill=blue!20,draw,font=\sffamily\Large\bfseries, minimum size=11mm, scale=0.5}]

\node[main node] (1) {};
\node[main node] (2) [below left of=1] {A};
\node[main node] (3) [below right of=1] {};
\node[main node] (4) [below left of=2] {B};
\node[main node] (5) [below right of=2] {C};
\node[main node] (6) [below right of=3] {D};

\path[every node/.style={font=\sffamily\small}]
    (1) edge node [left] {} (2)
	edge node [right] {} (3)
    (2) edge node [left] {} (4)
    edge node [right] {} (5)
	(3) edge node [right]{}(6);
	
	\node[draw=blue,inner sep=0pt,thick,fit=(4)] {};
		\node[draw=blue,inner sep=0pt,thick,fit=(5)] {};
	\node[draw=red,inner sep=0pt,circle,thick,fit=(2)] {};
		\node[draw=red,inner sep=0pt,circle,thick,fit=(6)] {};
\end{tikzpicture}\label{fig:ph4}}
\subfigure[Long distance problem]{
\begin{tikzpicture}[-,>=stealth',shorten >=1pt,auto,node distance=3cm,
  thick,main node/.style={circle,fill=blue!20,draw,font=\sffamily\Large\bfseries, minimum size=11mm, scale=0.48}]

\node[main node] (1) {};
\node[main node] (2) [below left of=1] {};
\node[main node] (3) [below right of=1] {};
\node[main node] (4) [below left of=2] {A};
\node[main node] (5) [below right of=2] {};
\node[main node] (6) [below right of=3] {B};

\path[every node/.style={font=\sffamily\small}]
    (1) edge node [left] {} (2)
	edge node [right] {} (3)
    (2) edge node [left] {} (4)
    edge node [right] {} (5)
	(3) edge node [right]{}(6);
	
	\node[draw=red,inner sep=0pt,thick,circle,fit=(4)] {};
		\node[draw=blue,inner sep=0pt,thick,fit=(6)] {};
\end{tikzpicture}\label{fig:ph5}}
\caption{Interesting cases when evaluating hierarchical classifiers. Nodes surrounded by circles are the true classes while the nodes surrounded by rectangles are the predicted classes.}
\label{phenomena1}
\end{figure}
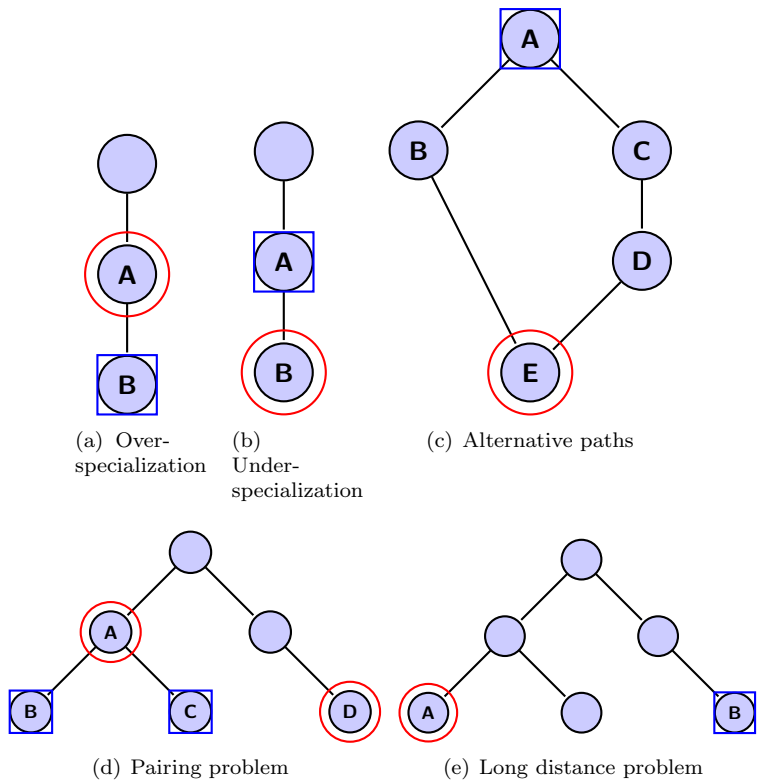

In order to measure the severity of an error in hierarchical classification, there are several interesting issues that need to be addressed. 
Figure \ref{phenomena1} presents five cases that require special handling. In all cases, the nodes surrounded by circles are the true classes, while the nodes surrounded by rectangles are the predicted ones. These cases can be sub-grouped in a) \emph{pairing} problems (Figures \ref{fig:ph4} and \ref{fig:ph5})
where one must select which pairs of predicted and true classes to take into account for the calculation of the error, and b) \emph{distance-measuring}
problems (Figures \ref{fig:ph1}, \ref{fig:ph2} and \ref{fig:ph3}) which concern the way that the error will be calculated for a pair of predicted and true classes.

Figure \ref{fig:ph2} presents an \emph{over-specialization} error where the predicted class
is a descendant of the true class. Figure \ref{fig:ph3} depicts an \emph{under-specialization} error, where an ancestor 
of the true class is selected. In both these cases the desired
behavior of the measure would be to reduce the penalty  to the classification system, according to the
distance between the true class and the predicted one.

The third case (Figure \ref{fig:ph1}), called \emph{alternative paths}, presents a scenario where there are two different ways to reach
the true class starting from a predicted class. In this case, a measure could use one of the 
two paths or both in order to evaluate the performance of the classification system.
Selecting the path that minimizes the distance between the two classes and using that as a measure of error seems
reasonable. In Figure \ref{fig:ph1} the predicted class is an ancestor of the true class, but an alternative paths case may also involve multiple paths
from an ancestor to a descendant predicted class.


Figure \ref{fig:ph4} presents a scenario which is common in multi-label
data. In this case one must decide, before even measuring the error, which pairs of true and 
predicted classes should be compared. For example, node \texttt{A} (true class) could be compared to \texttt{B} (predicted) and  \texttt{D} to \texttt{C}; 
or node \texttt{A} could be compared to both \texttt{B} and \texttt{C}, and node \texttt{D} to none;
other pairings are also possible. Depending on the pairings, the score assigned to the classifier will be different. It seems
reasonable to use the pairings that minimize the classification error. For example, in Figure \ref{fig:ph4} it could be argued that the prediction of \texttt{B} and
\texttt{C} are based on evidence about \texttt{A} and thus both \texttt{B} and \texttt{C} should be compared to A.

Finally, Figure \ref{fig:ph5} presents a case where the predicted class should probably not be matched
to any true class. This is typically the case when the predicted class and the true class
are too distant which is why we call this case the \emph{long distance problem}.

\subsection{Pair-based Measures}
Pair-based measures assign costs to pairs of predicted and
true classes. For example, in Figure \ref{fig:ph4} class \texttt{B} could be paired with
\texttt{A} and class \texttt{C} with \texttt{D}, and then the sum of the corresponding costs
would give the total misclassification error.

Let $\hat{Y}=\{\hat{y_i}|i=1\ldots M\}$ and $Y=\{y_j|j=1\ldots N\}$ be the sets of the predicted and true classes respectively, for a single test instance (the index of the instance
is omitted due to simplicity). The sets $Y$ and $\hat{Y}$ are augmented with a default predicted and a default true class, respectively corresponding to $\hat{y}_{M+1}$ and $y_{N+1}$. These classes are used when a predicted class cannot or should not be paired to any true class and vice-versa. For example, when the distances between a predicted class $\hat{y}_i$ and all the true classes $y_i$
exceed a predefined threshold (see the long distance problem in Figure \ref{fig:ph5}), the predicted class $\hat{y}_i$ may be paired with the default true
class.

Additionally, let
$\kappa_{ij}$ be the cost of predicting class $\hat{y}_i$ instead of the true
class $y_j$. The matrix $\mathbf{K}=[\kappa_{ij}]_{i=1\ldots M+1, j=1 \ldots N+1}$, $\kappa_{ij}\geq 0, \forall i, j$ contains the costs of all possible pairs of
 predicted and true classes, including the default classes.

Pair-based measures typically calculate the cost $\kappa_{ij}$ of a
pair of a predicted class $\hat{y}_i$ and a true class $y_i$ as the minimum distance of $\hat{y}_i$ and $y_i$ 
in the hierarchy, e.g. as the number of edges between the classes along the shortest path that connects them. The intuition is that the closer the two classes are in the hierarchy, the more similar they are, and therefore the less severe the error. More elaborate cost measures may assign weights to the hierarchy's edges, and the weights may decrease when moving from the top to the bottom \citep{Blockeel02,Holden06}. The distance to the default classes is usually set to a 
fixed large value.

In a spirit of fairness (minimum penalty), the aim of an evaluation measure is to pair the classes returned by a system and the true classes in a way that minimizes the overall classification error. This can be formulated as the following optimization problem:
\begin{myproblem}
\begin{eqnarray*}
\left\{
\begin{array}{ll}
 \displaystyle\min_{\substack{x_{ij},\\ 1 \le i \le M+1,\\ 1 \le j \le N+1}} \,\,\, \sum_{\substack{i =1 \ldots (N+1),\\ j=1\ldots (M+1)}} \kappa_{ij} x_{ij} & \\
 \mbox{subject to:} & \\
 \hspace{1cm} \mbox{(i)} \,\,\, \forall i=1\ldots M, \forall j=1\ldots N, x_{ij} \in \{0;1\}; \,\, x_{(M+1)(N+1)} = 0 & \\
 \hspace{1cm} \mbox{(ii)} \,\, \alpha_p \leq  \sum_{j=1}^{N+1} x_{ij} \leq \beta_p, \forall i=1\ldots M & \\
  \hspace{1cm} \mbox{(iii)} \,\, \alpha_t\leq \sum_{i=1}^{M+1} x_{ij}\leq \beta_t, \forall j=1\ldots N & \\
 \end{array}
\right.
\end{eqnarray*}
\label{problem1}
\end{myproblem}

Constraint (i) states that $x_{ij}$, which denotes the alignment between classes, is either 0 (classes $\hat{y}_i$ and $y_j$ are not paired) or 1 (classes $\hat{y}_i$ and $y_j$ are paired); it furthermore states that the default predicted and true classes cannot be aligned (these default classes are solely used to ``collect" those predicted and true classes with no counterpart). The parameters $\alpha_p$, $\beta_p  \in  \mathbb{N}$ (constraint (ii)) are the lower and upper bounds of the allowed number of true classes that a predicted class can be paired with.  For example, setting $\alpha_p=\beta_p=1$  requires each predicted class to be paired with exactly one true class. Similarly, the parameters $\alpha_t$, $\beta_t  \in  \mathbb{N}$ (constraint (iii)) limit the number of predicted labels that a true class can be paired with. The above constraints directly imply\footnote{\label{foot:lab}Indeed, in the worst case, i.e. when all $x_{ij}$ but $x_{iN+1}$ are 0, constraint (ii) yields $x_{iN+1} = \beta_p$; 
the reasoning is similar for constraint (iii).}  that $\forall i=1\ldots M, x_{iN+1} \leq \beta_p$ and $\forall j=1\ldots N,  x_{M+1j} \leq \beta_t$, meaning that the default true class can be aligned to at most $\beta_pM$ predicted classes and the default predicted class to at most $\beta_tN$ true classes.


The above problem corresponds to a best pairing problem in a bipartite graph, the nodes of which being respectively the predicted and true classes. It is important to note here that the pairing we are looking for is not a matching, since the same node can be paired with several nodes.
We opt to approach this problem as a graph pairing one rather than a linear optimization one, for two reasons: first because there exist polynomial solutions to pairing problems in graphs, and second because the graph framework allows one to easily illustrate how the different cost-based measures proposed so far relate to each other.
In particular, we model it as a cost flow minimization problem \citep{Ahuja93}.
\subsubsection{A Flow Network Model for Class Pairing}

A flow network is a directed graph $G=(V,E)$ with $m$ edges, where each edge $u$ $\in E$ is associated
with a lower and an upper capacity denoted $b_u$ and $c_u$ respectively. The flow along an
edge $u$ is denoted as $\phi_u$ and $b_u\leq \phi_u \leq c_u$.
The flow of the network is a vector $\phi = (\phi_1, \phi_2,\ldots,\phi_m)^T$ $\in \mathbb{R}^m$. For each vertex $i \in V$,
the flow conservation property holds:
\[
\sum_{u\in \omega^{+}(i)} \phi_u = \sum_{u\in \omega^{-}(i)} \phi_u
\]
where $\omega^{+}(i)$ and $\omega^{-}(i)$ denote the set of edges entering
and leaving vertex $i$ respectively. Each edge $u$ is also associated with a cost $\gamma_u$ which represents
the cost of using this edge. The total cost of a flow $\phi$ is:
\[
\gamma^T \times \phi = \sum_{u\in E}\gamma_u  \phi_u,
\]
where $\gamma=(\gamma_1, \gamma_2,\ldots,\gamma_m)^T$ $\in \mathbb{R}^m$. The minimum cost flow is the one that minimizes $\gamma^T \phi$ while satisfying the capacity and flow conservation constraints. The quantity to be minimized in flow networks is the same as the one in Problem~\ref{problem1}, the constraints in this latter problem corresponding to capacity constraints, as explained below. Furthermore, the following integrality theorem states that when the bounds of capacity intervals are integers, there exists a minimal cost flow such that the quantity of flow on each edge is also an integer:
\begin{mytheorem}
If a flow network has capacities which are all integer valued and there exists some feasible flow in the network, then there is a minimum cost feasible flow with an integer valued flow on every arc.
\end{mytheorem}
Furthermore, all standard algorithms for finding minimal cost flows guarantee to find this particular flow \citep{Ahuja93}.

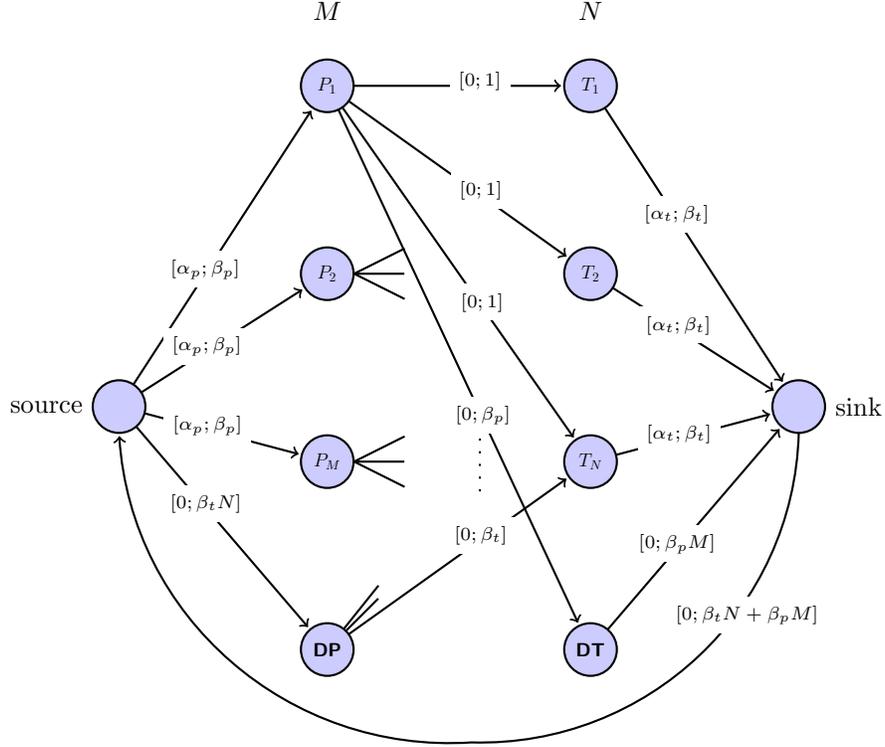
\begin{figure}[ht]
\begin {center}
\begin{tikzpicture}[bend angle=120, shorten >=1pt,auto,node distance=5cm, 
  thick,main node/.style={circle,fill=blue!20,draw,font=\sffamily\Large\bfseries, minimum size=14mm, scale=0.5}
  ]

\node[main node] (1) {$P_1$};
\node[main node] (2) [right of=1, xshift=20mm] {$T_1$};
\node[main node] (3) [below of =1] {$P_2$};
\node[main node] (4) [below of =2] {$T_2$};
\node[main node] (5) [below of =3] {$P_M$};
\node[main node] (6) [below of =4] {$T_N$};
\node[main node] (7) [below of =5] {DP};
\node[main node] (8) [below of =6] {DT};
\node[main node] (9) [below left of =3,xshift=-20mm, label= left:source] {};
\node[main node] (10) [below right of =4,xshift=20mm, label= right:sink] {};

\node[draw=none, node distance=1cm](mlabel)[above of=1]{$M$};
\node[draw=none, node distance=1cm](nlabel)[above of=2]{$N$};

\path[every node/.style={font=\sffamily\small},->]
    (1) edge node(l1) [pos=0.6, anchor=base, fill=white] {\scriptsize $[0;1]$} (2)	
	edge node [pos=0.6, anchor=base, fill=white] {\scriptsize $[0;\beta_p]$} (8)
	  edge node(l2) [pos=0.6, anchor=base, fill=white] {\scriptsize $[0;1]$} (4)
	  edge node(l3) [pos=0.6, anchor=base, fill=white] {\scriptsize $[0;1]$} (6)

    (2) edge node [pos=0.4, anchor=base, fill=white] {\scriptsize $[\alpha_t;\beta_t]$} (10)
    (4) edge node [pos=0.4, anchor=base, fill=white] {\scriptsize $[\alpha_t;\beta_t]$} (10)
    (6) edge node [pos=0.4, anchor=base, fill=white] {\scriptsize $[\alpha_t;\beta_t]$} (10)
    (8) edge node [pos=0.4, anchor=base, fill=white] {\scriptsize $[0;\beta_p M]$} (10)
    (9) edge node [pos=0.4, anchor=base, fill=white] {\scriptsize $[\alpha_p;\beta_p]$} (1)
	edge node [pos=0.4, anchor=base, fill=white] {\scriptsize $[\alpha_p;\beta_p]$} (3)
	edge node [pos=0.4, anchor=base, fill=white] {\scriptsize $[\alpha_p;\beta_p]$} (5)
	edge node [pos=0.4, anchor=base, fill=white] {\scriptsize $[0;\beta_tN]$} (7)
	(7) edge node [pos=0.6, anchor=base, fill=white] {\scriptsize $[0;\beta_t]$} (6)	
	;
    \draw[] (3.east) --+ (20pt,0);
    \draw[] (3.east) --+ (20pt,-10pt);
    \draw[] (3.east) --+ (20pt,10pt);
       \draw[] (5.east) --+ (20pt,0);
        \draw[] ( 5.east) --+ (20pt,-10pt);
        \draw[] (5.east) --+ (20pt,10pt);
                \draw[loosely dotted] ($(l1.south)+(0,-4.5cm)$) -- +(0,-25pt);
\draw[->] (10.south) to[bend left=45]node[pos=0.4,anchor=base,fill=white,xshift=0.2cm]{\scriptsize $[0;\beta_t N+\beta_p M]$} ($(8.south)+(-45pt,-25pt)$) to[bend left=45]   (9.south); 
    \draw[] (7) --+ (20pt,20pt);
        \draw[] (7) --+ (20pt,25pt);
\end{tikzpicture}
\caption{The proposed flow network.}
\label {fig:flownetwork}
\end {center}
\end{figure}

Pairing problems in bipartite graphs are represented with flow networks by adding two nodes, a source and a sink, and edges from the source to the first set of nodes, from the second set of nodes to the sink, and from the sink to the source. These extra nodes and edges ensure that the flow conservation constraints are satisfied.
For pair-based measures, one thus obtains the following flow network framework $G(V,E)$ (see also Figure \ref{fig:flownetwork}):
\begin{itemize}
\item $V$ includes a source, a sink, the predicted classes, the true classes, a default true class and a default predicted class;
\item $E$ includes edges from the source to all the predicted classes (including the default predicted class), from every predicted class to every true class (including the default true class), from every true class to the sink and from the sink to the source.
\end{itemize}
No edges exist between the default predicted and default true class, as required by constraint (i) above.

In our setting the capacity interval $[b_u;c_u]$ of an edge $u$  expresses the possible number of pairs that each predicted or true class can participate in. 
The interval between each pair of predicted and true classes restricts the flow on that network which indicates whether this pair will be considered in the calculation of the evaluation measure. Put it differently, in the solved flow network the flow values will reflect the specific evaluation measure as they show the pairs that make up the solution with the minimum cost. The intervals between the source and the predicted classes as well as between the true classes and the sink also affect the way that the pairing will
be performed.

Due to the constraints in Problem~\ref{problem1}, the capacity intervals are defined as follows:
\begin{itemize}
\item From each predicted class $\hat{y_j}$ to each true class $y_i$, excluding the default class, the capacity interval is [0;1]; the integrality theorem here implies that the flow value between predicted and true classes will be either 0 or 1, i.e. a predicted and a true class either be paired (1) or not paired (0). The capacity bounds here correspond to the $x_{ij}$ values of Problem~\ref{problem1} (constraint (i));
\item From the source to a (non-default) predicted class, the capacity interval is $[\alpha_p;\beta_p]$ meaning that a predicted class is aligned with at least $\alpha_p$ (and at most $\beta_p$) true classes;
\item Similarly, from a (non-default) true class to the sink, the capacity intervals is [$\alpha_t$;$\beta_t$] meaning that a true class is aligned with at least $\alpha_t$ (and at most $\beta_t$) predicted classes;
\item From each predicted class $\hat{y_j}$ to the default true class the capacity interval is [0;$\beta_p$] and from the default predicted class to each true class the capacity interval is [0;$\beta_t$]; from the source (resp. sink) to the default predicted (resp. true) class, the capacity interval is $[0;\beta_t N]$ (resp. $[0;\beta_pM]$), as mentioned in footnote~\ref{foot:lab};
\item Lastly, from the sink to the source, the capacity interval is [0;$\beta_t N+\beta_p M$], which corresponds to a loose setting compatible with the intervals given above (this last capacity interval does not impose any constraint but is necessary to ensure flow conservation).
\end{itemize}


\subsubsection{Existing Pair-based Measures}
The majority of the existing pair-based measures deals only with tree hierarchies and single-label
problems. Under these conditions the pairing problem becomes simple, because a single
path exists between the predicted and the true classes. The complexity of the problem increases
when the hierarchy is a DAG or when the problem is multi-labeled; current measures cannot handle the majority
of the phenomena presented in Section \ref{sec:problems}.

In the simplest case of pair-based measures \citep{Dekel04,Holden06}, the measure trivially pairs the
single prediction with the single true label ($M=N=1$), so that $\alpha_p=\beta_p=\alpha_t=\beta_t=1$. Note that no default classes exist in this measure, or equivalently the corresponding costs are equal to infinity, $\gamma(DP,T)=\gamma(DT,P)=+\infty$

\begin{figure}[ht]
\begin {center}
\begin{tikzpicture}[->,>=stealth' ,bend angle=90, shorten >=1pt,auto,node distance=5cm, 
  thick,main node/.style={circle,fill=blue!20,draw,font=\sffamily\Large\bfseries, minimum size=14mm, scale=0.55}]

\node[main node] (1) {P};
\node[main node] (2) [right of=1] {T};
\node[main node] (3) [below of =1] {DP};
\node[main node] (4) [below of =2] {DT};
\node[main node] (5) [below left of =1, label= left:source] {};
\node[main node] (6) [below right of =2, label= right:sink] {};

\path[every node/.style={font=\sffamily\small}]
    (1) edge node [anchor=base, fill=white] {\scriptsize [0;1]} (2)	
	edge node [pos=0.3, anchor=base, fill=white] {\scriptsize [0;1]} (4)
    (2) edge node [anchor=base, fill=white] {\scriptsize [0;1]} (6)
    (3) edge node [pos=0.3, anchor=base, fill=white] {\scriptsize [0;1]} (2)		
    (4) edge node [anchor=base, fill=white] {\scriptsize [0;1]} (6)
    (5) edge node [anchor=base, fill=white] {\scriptsize [1;1]} (1)
	edge node [anchor=base, fill=white] {\scriptsize [0;1]} (3)
    (6) edge [bend left] node [anchor=base, fill=white] {\scriptsize [1;2]} (5);
\end{tikzpicture}
\caption{Tree Induced Error flow network}
\label {treeInducedError}
\end {center}
\end{figure}
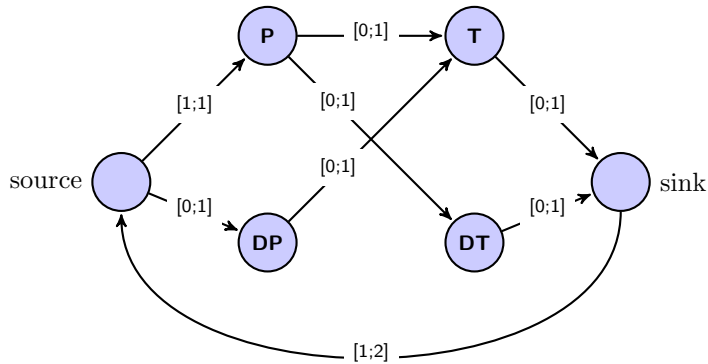

For a pair $(\hat{y_j},y_i)$, of a predicted and a true class, depicted as $P$ and $T$ respectively in Figure \ref{treeInducedError},
$\gamma(\hat{y_j},y_i)=\gamma(P,T)=\kappa_{ij}$ is taken to be the distance between $y_i$ and $\hat{y_j}$: 
\begin{equation}
\kappa_{ij}=\sum_{e \in E(i,j)}w_e,
\label{eq1}
\end{equation}
where $E(i,j)$ is the set of edges along the path from $y_i$ to $\hat{y_j}$ in the hierarchy and $w_e$ is the weight
of edge $e$. For $w_e=1$, we get what \cite{Dekel04} call \emph{tree induced error}.


In \citep{Sun01} two cost measures are proposed for multi-label problems in tree hierarchies, where all possible pairs of the predicted and true classes are used in the calculation. In this case, $\alpha_p=\beta_p=N$ and $\alpha_t=\beta_t=M$. Again, no default classes are used and so the corresponding costs are: $\gamma(DP,y_i)=\gamma(DT,\hat{y_j})=+\infty, \  i =1\ldots N, j=1\ldots M$. Note that this is an extreme case, where all pairs of predicted and true labels are used. The weights $w_e$ are calculated in two alternative ways: a) as the similarity (e.g., cosine similarity) between the classes of the predicted and true ones, and b) using the distances of the hierarchy as in Equation \ref{eq1}.

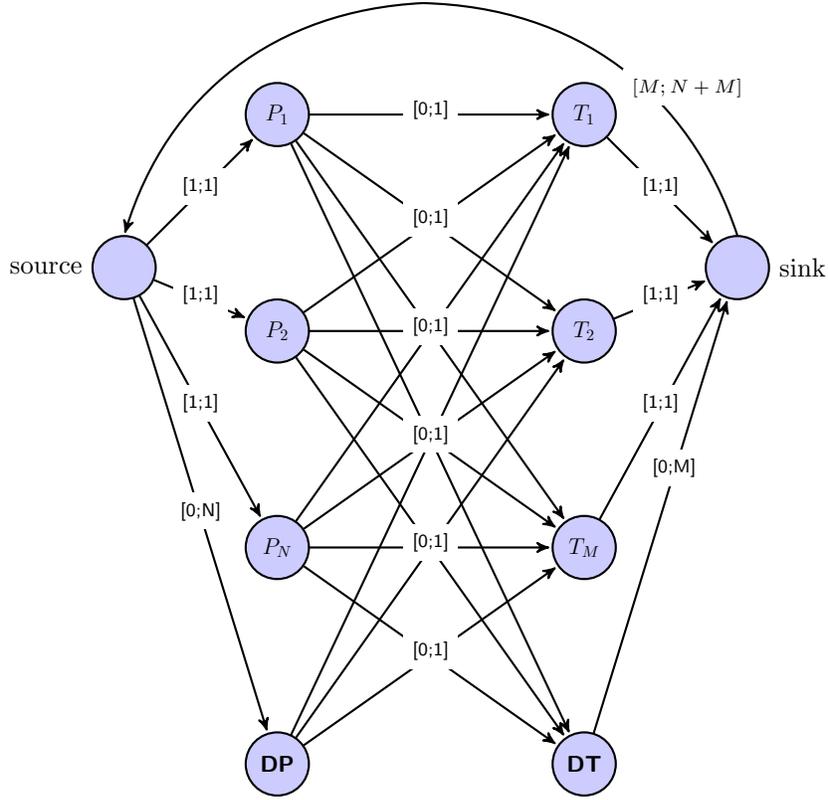
\begin{figure}[h]
\begin {center}
\begin{tikzpicture}[->,>=stealth' ,bend angle=120, shorten >=1pt,auto,node distance=4.8cm, 
  thick,main node/.style={circle,fill=blue!20,draw,font=\sffamily\Large\bfseries, minimum size=14mm, scale=0.6}]

\node[main node] (1) {$P_1$};
\node[main node] (2) [right of=1, xshift=20mm] {$T_1$};
\node[main node] (3) [below of =1] {$P_2$};
\node[main node] (4) [below of =2] {$T_2$};
\node[main node] (5) [below of =3] {$P_N$};
\node[main node] (6) [below of =4] {$T_M$};
\node[main node] (7) [below of =5] {DP};
\node[main node] (8) [below of =6] {DT};
\node[main node] (9) [below left of =1, label= left:source] {};
\node[main node] (10) [below right of =2, label= right:sink] {};

\path[every node/.style={font=\sffamily\small}]
    (1) edge node [anchor=base, fill=white] {\scriptsize [0;1]} (2)	
	edge node [anchor=base, fill=white] {\scriptsize [0;1]} (4)
	edge node [anchor=base, fill=white] {\scriptsize [0;1]} (6)
	edge node [anchor=base, fill=white] {\scriptsize [0;1]} (8)
    (3) edge node [anchor=base, fill=white] {\scriptsize [0;1]} (2)	
	edge node [anchor=base, fill=white] {\scriptsize [0;1]} (4)
	edge node [anchor=base, fill=white] {\scriptsize [0;1]} (6)
	edge node [anchor=base, fill=white] {\scriptsize [0;1]} (8)
    (5) edge node [anchor=base, fill=white] {\scriptsize [0;1]} (2)	
	edge node [anchor=base, fill=white] {\scriptsize [0;1]} (4)
	edge node [anchor=base, fill=white] {\scriptsize [0;1]} (6)
	edge node [anchor=base, fill=white] {\scriptsize [0;1]} (8)
    (7) edge node [anchor=base, fill=white] {\scriptsize [0;1]} (2)	
	edge node [anchor=base, fill=white] {\scriptsize [0;1]} (4)
	edge node [anchor=base, fill=white] {\scriptsize [0;1]} (6)	
    (2) edge node [anchor=base, fill=white] {\scriptsize [1;1]} (10)
    (4) edge node [anchor=base, fill=white] {\scriptsize [1;1]} (10)
    (6) edge node [anchor=base, fill=white] {\scriptsize [1;1]} (10)
    (8) edge node [pos=0.6, anchor=base, fill=white] {\scriptsize [0;M]} (10)
    (9) edge node [anchor=base, fill=white] {\scriptsize [1;1]} (1)
	edge node [anchor=base, fill=white] {\scriptsize [1;1]} (3)
	edge node [anchor=base, fill=white] {\scriptsize [1;1]} (5)
	edge node [anchor=base, fill=white] {\scriptsize [0;N]} (7);
\draw[->] (10.north) to[bend right=35]node[pos=0.4,anchor=base,fill=white,xshift=0.5cm]{\scriptsize $[ M; N+ M]$} ($(1.north)+(55pt,30pt)$) to[bend right=35]   (9.north); 
\end{tikzpicture}
\caption{Graph Induced Error (GIE) flow network.}
\label {GIEFlow}
\end {center}
\end{figure}

A measure dubbed Graph Induced Error (GIE) was proposed and used during the second
Large Scale Hierarchical Text classification challenge (LSHTC)\footnote{http://lshtc.iit.demokritos.gr/}. GIE
is based on the best matching pairs of predicted and true classes and can handle multi-label (and single-labeled)
classification with both tree and DAG class hierarchies. For a particular instance being classified, each predicted class
is paired either with one true class or with the default true class; multiple predicted classes can be paired with
the default true class (Figure \ref{GIEFlow}). Similarly, each true class is paired with exactly one predicted class
or with the default predicted class, and several true classes can be paired with the default predicted class.
Hence,  $\alpha_p=\beta_p=\alpha_t=\beta_t=1$. The cost $\kappa_{ij}$ is computed as in Equation \ref{eq1},
with $w_e=1, \forall e$. If the hierarchy is a DAG, multiple hierarchy paths may link each predicted class $\hat{y_j}$
to its paired true class $y_i$; then $E(i,j)$ is taken to be the shortest of these paths.
The cost of pairing a class (predicted or true) with a default one is set to a positive value $D_{\max}$.
Figure \ref{GIEFlow} presents the corresponding flow network.

In multi-label classification GIE's concept  of ``best'' matching  fails to address 
the pairing problem of Section \ref{sec:problems}. For example, if a predicted class has two true classes as children, as in Figure \ref{fig:ph4}, then only one of them would be paired with its parent. The other one would either be penalized with $D_{\max}$ or would be paired with another distant class.

\subsubsection{Multi-label Graph Induced Accuracy}

\begin{figure}[h]
\begin {center}
\begin{tikzpicture}[->,>=stealth' ,bend angle=120, shorten >=1pt,auto,node distance=4.8cm, 
  thick,main node/.style={circle,fill=blue!20,draw,font=\sffamily\Large\bfseries, minimum size=14mm, scale=0.6}]

\node[main node] (1) {$P_1$};
\node[main node] (2) [right of=1, xshift=20mm] {$T_1$};
\node[main node] (3) [below of =1] {$P_2$};
\node[main node] (4) [below of =2] {$T_2$};
\node[main node] (5) [below of =3] {$P_M$};
\node[main node] (6) [below of =4] {$T_N$};
\node[main node] (7) [below of =5] {DP};
\node[main node] (8) [below of =6] {DT};
\node[main node] (9) [below left of =1, label= left:source] {};
\node[main node] (10) [below right of =2, label= right:sink] {};

\path[every node/.style={font=\sffamily\small}]
    (1) edge node [anchor=base, fill=white] {\scriptsize [0;1]} (2)	
	edge node [anchor=base, fill=white] {\scriptsize [0;1]} (4)
	edge node [anchor=base, fill=white] {\scriptsize [0;1]} (6)
	edge node [anchor=base, fill=white] {\scriptsize [0;1]} (8)
    (3) edge node [anchor=base, fill=white] {\scriptsize [0;1]} (2)	
	edge node [anchor=base, fill=white] {\scriptsize [0;1]} (4)
	edge node [anchor=base, fill=white] {\scriptsize [0;1]} (6)
	edge node [anchor=base, fill=white] {\scriptsize [0;1]} (8)
    (5) edge node [anchor=base, fill=white] {\scriptsize [0;1]} (2)	
	edge node [anchor=base, fill=white] {\scriptsize [0;1]} (4)
	edge node [anchor=base, fill=white] {\scriptsize [0;1]} (6)
	edge node [anchor=base, fill=white] {\scriptsize [0;1]} (8)
    (7) edge node [anchor=base, fill=white] {\scriptsize [0;1]} (2)	
	edge node [anchor=base, fill=white] {\scriptsize [0;1]} (4)
	edge node [anchor=base, fill=white] {\scriptsize [0;1]} (6)	
    (2) edge node [anchor=base, fill=white] {\scriptsize [1;M]} (10)
    (4) edge node [anchor=base, fill=white] {\scriptsize [1;M]} (10)
    (6) edge node [anchor=base, fill=white] {\scriptsize [1;M]} (10)
    (8) edge node [anchor=base, fill=white] {\scriptsize [0;M]} (10)
    (9) edge node [anchor=base, fill=white] {\scriptsize [1;N]} (1)
	edge node [anchor=base, fill=white] {\scriptsize [1;N]} (3)
	edge node [anchor=base, fill=white] {\scriptsize [1;N]} (5)
	edge node [anchor=base, fill=white] {\scriptsize [0;N]} (7);
    \draw[->] (10.north) to[bend right=35]node[pos=0.4,anchor=base,fill=white,xshift=0.7cm]{\scriptsize $[ M; M(N+ 1)]$} ($(1.north)+(55pt,30pt)$) to[bend right=35]   (9.north); 
\end{tikzpicture}
\caption{Multi-label Graph Induced Accuracy flow network.}
\label {ISBMFlow}
\end {center}
\end{figure}
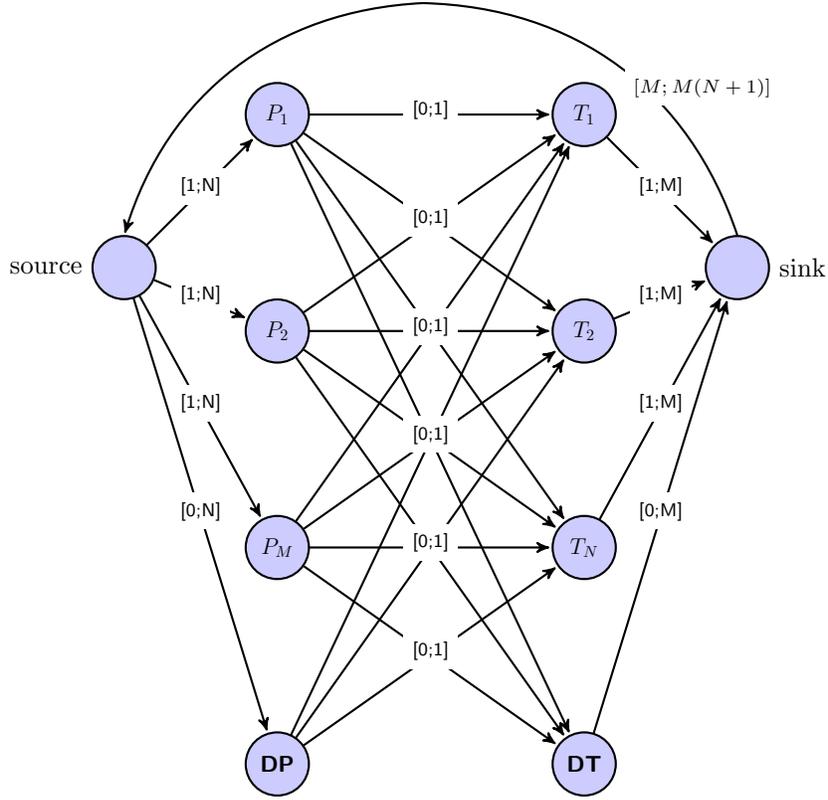

We propose here a straightforward extension of GIE called Multi-label Graph Induced Accuracy (MGIA), in which each class is allowed to participate in more than one pair. This extension makes the 
method more suitable to the pairing problem. Figure \ref{ISBMFlow} presents the  MGIA flow network, 
in which $\alpha_p=\alpha_t=1$, $\beta_p=N$, $\beta_t=M$. The cost  of pairing a class (predicted or true) with a default one is set as in GIE. 
Solving the flow network optimization problem is easy since the only constraints are that the default predicted class cannot be paired with the default true 
class and that categories of the same set (predicted or true) cannot be paired to each other. Thus each  pairing can be solved separately from the others
 by pairing a class with either the default class of the other set, or the nearest class of the other set.

As in the previous pair-based measures, after the solution of the problem an error is calculated on the solved network. Instead of using directly this error for evaluation we define an accuracy based measure as follows:
\[
1 - \frac{fnerror}{|P\cup T \setminus P\cap T|*D_{max}}
\]
where $fnerror$ is the value provided by the solved flow network.

The above measure is bounded in [0,1] and the better a system is the closer it will be to 1. Note that
in the case where all predicted classes and all true classes are paired with the respective default classes,
$fnerror$ will reach its maximum value $|P\cup T|*D_{max}$ and will be equal to the denominator as
$P\cap T =\emptyset$ resulting in a value of 0. Essentially, the advantage of the proposed measure
over other pair-based measures is that it takes into account the correct predictions of the classification system (that is the true positives, $P \cap T$).

\subsection{Set-based Measures}
The performance measures of this category are based on operations on the entire sets of predicted and true classes,
possibly including also their ancestors and descendants, as opposed to pair-based measures, which consider only pairs
of predicted and true classes.

Set-based measures have two distinct phases:
\begin{enumerate}
\item The augmentation of $Y$ and $\hat{Y}$ with information about the hierarchy.
\item The calculation of a cost measure based on the augmented sets.
\end{enumerate}

The augmentation of $Y$ and $\hat{Y}$ is a crucial step, attempting to capture
the hierarchical relations of the classes. For example, the sets may be
augmented with the ancestors of the true and predicted classes as follows:
\begin{equation}
Y_{aug}=Y \cup An(y_1) \cup \ldots \cup An(y_N)
\label{eq2}
\end{equation}
\begin{equation}
\hat{Y}_{aug}=\hat{Y}\cup An(\hat{y}_1) \cup \ldots \cup An(\hat{y}_M)
\label{eq3}
\end{equation}
Using the augmented sets of predicted and true classes, two approaches have mainly been adopted 
to calculate the misclassification cost: a) symmetric difference loss and b) hierarchical precision and recall.

Symmetric difference loss is calculated as follows, where $|S|$ the cardinality of a set $S$:
\[
l_{\Delta}(Y_{aug},\hat{Y}_{aug}) = |(\hat{Y}_{aug} \setminus Y_{aug})\cup (Y_{aug} \setminus \hat{Y}_{aug})|
\]
If we use the initial $\hat{Y}$ and $Y$ sets instead of $\hat{Y}_{aug}, Y_{aug}$ , the measure becomes the standard symmetric difference
for flat multi-label classification. Also, note that the two quantities of the symmetric loss difference express the false positive and false negative
rates respectively.

On the other hand, hierarchical precision and recall are defined as follows:

\[
P_{H} = \frac{|\hat{Y}_{aug}\cap Y_{aug}|}{|\hat{Y}_{aug}|}
\]
\[
R_{H} = \frac{|\hat{Y}_{aug}\cap Y_{aug}|}{|Y_{aug}|}
\]
The nominator of these measures expresses the true positive rate and can be written as follows:
\[
|\hat{Y}_{aug}\cap Y_{aug}| = |\hat{Y}_{aug}\cup Y_{aug}| - l_{\Delta}((Y_{aug},\hat{Y}_{aug})
\]
where we note that the symmetric loss is a substractive term.

Set-based measures are not affected by the pairing problem of Figure \ref{fig:ph4} and the long distance problem of Figure \ref{fig:ph5}, as they do not
rely on pairing of true and predicted classes.
\subsubsection{Existing Set-based Measures}

Different measures differ mainly in the way the sets of predicted and true classes are augmented.
In \citep{Kiritchenko05,Struyf05,Cai07} the ancestors of the predicted and true classes are added to 
$Y_{aug}$ and $\hat{Y}_{aug}$, as in Equations \ref{eq2} and \ref{eq3} above.
Alternatively, in \cite{Ipeirotis01} the descendants of the true and predicted
classes are added:
\[
Y_{aug}=Y \cup De(y_1) \cup \ldots \cup De(y_N)
\]
\[
\hat{Y}_{aug}=\hat{Y}\cup De(\hat{y}_1) \cup \ldots \cup De(\hat{y}_M)
\]
In the latter approach, when the true and predicted classes are in
different sub-graphs of the hierarchy (different sub-trees, if the hierarchy is a tree), a maximum penalty will be given, even when
several ancestors have been correctly predicted.

In \citep{Bianchi06}, the approach that  adds the ancestors is adopted (Equations 2 and 3) but the augmented sets are then altered as
follows:

\begin{equation}
Y_{aug} \leftarrow Y_{aug}\setminus \left\lbrace y_k \biggm\vert  
y_k \in Y_{aug} \wedge y_k \notin \hat{Y}_{aug}\wedge Pa(y_k)\notin  \hat{Y}_{aug}\right\rbrace
\label{eq4}
\end{equation}

\begin{equation}
\hat{Y}_{aug} \leftarrow \hat{Y}_{aug}\setminus \left\lbrace \hat{y_k} \biggm\vert 
\hat{y_k} \in \hat{Y}_{aug} \wedge \hat{y_k} \notin Y_{aug}\wedge Pa(\hat{y_k})\notin  Y_{aug} \right\rbrace
\label{eq5}
\end{equation}

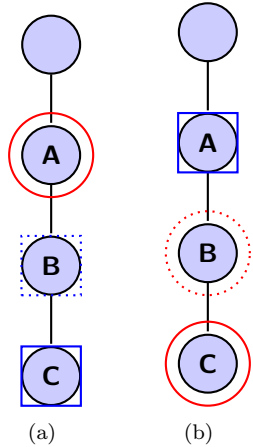
\begin{figure}[H]
\centering
\subfigure[]{
\begin{tikzpicture}[-,>=stealth',shorten >=1pt,auto,node distance=3cm,
  thick,main node/.style={circle,fill=blue!20,draw,font=\sffamily\Large\bfseries, minimum size=11mm, scale=0.7}]

\node[main node] (1) {};
\node[main node] (2) [below of=1,,yshift=9mm] {A};
\node[main node] (3) [below of=2,,yshift=9mm] {B};
\node[main node] (4) [below of=3,,yshift=9mm] {C};

\path[every node/.style={font=\sffamily\small}]
    (1) edge node [] {} (2)
    (2) edge node [right] {} (3)
    (3) edge node[]{}(4);

\node[draw=red,inner sep=0pt,thick,circle,fit=(2)] {};
\node[draw=blue,dotted,inner sep=0pt,thick,fit=(3)] {};
\node[draw=blue,inner sep=0pt,thick,fit=(4)] {};
\end{tikzpicture}\label{fig:ex1}}
\qquad
\subfigure[]{
\begin{tikzpicture}[-,>=stealth',shorten >=1pt,auto,node distance=3cm,
  thick,main node/.style={circle,fill=blue!20,draw,font=\sffamily\Large\bfseries, minimum size=11mm, scale=0.7}]

\node[main node] (1) {};
\node[main node] (2) [below of=1,,yshift=9mm] {A};
\node[main node] (3) [below of=2,,yshift=9mm] {B};
\node[main node] (4) [below of=3,,yshift=9mm] {C};

\path[every node/.style={font=\sffamily\small}]
    (1) edge node [] {} (2)
    (2) edge node [right] {} (3)
    (3) edge node []{}(4);

\node[draw=red,dotted,circle,inner sep=0pt,thick,fit=(3)] {};
\node[draw=blue,inner sep=0pt,thick,fit=(2)] {};
\node[draw=red,inner sep=0pt,thick,circle,fit=(4)] {};
\end{tikzpicture}\label{fig:ex2}}
\caption{Tolerating over-specialization and under-specialization.}
\label{fig:examples}
\end{figure}

Equation \ref{eq5} introduces some tolerance to over-specialization. Consider, for example,
Figure \ref{fig:ex1} where we assume that the only true class is A and the only predicted class is C. According to Equation \ref{eq3}
we add class B (and class A) to $\hat{Y}_{aug}$. Based on Equation \ref{eq5} we then remove C from $\hat{Y}_{aug}$ to avoid penalizing
the classification method for B and C. Similarly, with Equation \ref{eq4} we tolerate  under-classification. In Figure \ref{fig:ex2}
the only true class is C and the only predicted class is A. According to Equation \ref{eq2} class B (and A) are added to $Y_{aug}$. 
Based on Equation \ref{eq4} then we remove C from $\hat{Y}_{aug}$ to avoid penalizing the classification method for both B and C.
The drawback of this measure is that it tends to favor category systems that stop their predictions early in the hierarchy.

\subsubsection{Lowest Common Ancestor Precision, Recall and $F_1$ Measures}

The approach proposed in this paper is based on the hierarchical versions of precision, recall and $F_1$, which add all the ancestors of the predicted and true classes to $Y_{aug}$ and $\hat{Y}_{aug}$.
Adding all the ancestors has the undesirable effect of over-penalizing errors that happen to nodes with many ancestors. 

In an attempt to address this issue, we propose the Lowest Common Ancestor Precision ($P_{LCA}$), Recall ($R_{LCA}$) and $F_1$ ($F_{LCA}$) measures. 
These measures use the concept of the lowest common ancestor ($LCA$) as defined in graph theory  \citep{Aho73}. 
\newtheorem{LCA}{Definition}
\begin{LCA}
The lowest common ancestor $LCA(n_1,n_2)$ of two nodes $n_1$ and $n_2$ of a tree $T$ is defined as the lowest node in $T$ (furthest from the root) that is an ancestor of both $n_1$ and $n_2$. 

\begin{figure}[ht]

\subfigure[Tree] {\label{Tree} 
\begin{tikzpicture}[-,>=stealth',shorten >=1pt,auto,node distance=3cm, 
  thick,main node/.style={circle,fill=blue!20,draw,font=\sffamily\Large\bfseries, minimum size=14mm, scale=0.4}]

\node[main node] (1) {0};
\node[main node] (2) [below left of=1, xshift=-20mm] {1};
\node[main node] (3) [below of=1, yshift=9mm] {2};
\node[main node] (4) [below right of=1, xshift=20mm] {3};
\node[main node] (5) [below left of=2] {1.1};
\node[main node] (6) [below right of=2] {1.2};
\node[main node] (7) [below of=3, yshift=9mm] {2.1};
\node[main node] (8) [below left of=4] {3.1};
\node[main node] (9) [below of=4, yshift=9mm] {3.2};
\node[main node] (10) [below right of=4] {3.3};
\node[main node] (11) [below left of=9] {3.2.1};
\node[main node] (12) [below right of=9] {3.2.2};

\path[every node/.style={font=\sffamily\small}]
    (1) edge node [left] {} (2)
	edge node [] {} (3)
	edge node [right] {} (4)
    (2) edge node [left] {} (5)
	edge node [right] {} (6)
    (3) edge node [] {} (7)
    (4) edge node [left] {} (8)
	edge node [] {} (9)
	edge node [right] {} (10)
    (9) edge node [left] {} (11)
	edge node [right] {} (12);
\end{tikzpicture}
}
\subfigure[DAG] {\label{DAG}
\begin{tikzpicture}[-,>=stealth',shorten >=1pt,auto,node distance=3cm, 
  thick,main node/.style={circle,fill=blue!20,draw,font=\sffamily\Large\bfseries, minimum size=14mm, scale=0.4}]

\node[main node] (1) {0};
\node[main node] (2) [below left of=1, xshift=-20mm] {1};
\node[main node] (3) [below of=1, yshift=9mm] {2};
\node[main node] (4) [below right of=1, xshift=20mm] {3};
\node[main node] (5) [below left of=2] {1.1};
\node[main node] (6) [below right of=2] {1.2};
\node[main node] (7) [below of=3, yshift=9mm] {2.1};
\node[main node] (8) [below left of=4] {3.1};
\node[main node] (9) [below of=4, yshift=9mm] {3.2};
\node[main node] (10) [below right of=4] {3.3};
\node[main node] (11) [below left of=9] {3.2.1};
\node[main node] (12) [below right of=9] {3.2.2};

\path[every node/.style={font=\sffamily\small}]
    (1) edge node [left] {} (2)
	edge node [] {} (3)
	edge node [right] {} (4)
    (2) edge node [left] {} (5)
	edge node [right] {} (6)
    (3) edge node [] {} (7)
    (4) edge node [left] {} (8)
	edge node [left] {} (7)
	edge node [] {} (9)
	edge node [right] {} (10)
	edge node [right] {} (12)
    (9) edge node [left] {} (11)
	edge node [right] {} (12);

\node[draw=blue,inner sep=0pt,thick,fit=(8)] {};
\node[draw=blue,inner sep=0pt,thick,fit=(11)] {};
\node[draw=blue,inner sep=0pt,thick,fit=(12)] {};
\node[draw=red,inner sep=0pt,thick,circle,fit=(7)] {};
\node[draw=red,inner sep=0pt,thick,circle,fit=(10)] {};
\node[draw=red,inner sep=0pt,thick,circle,fit=(11)] {};
\end{tikzpicture}
}
\caption{Tree and DAG hierarchies. Nodes surrounded by circles are true classes. Nodes surrounded by rectangles are predicted classes.}
\label {treeDAG}
\end{figure}
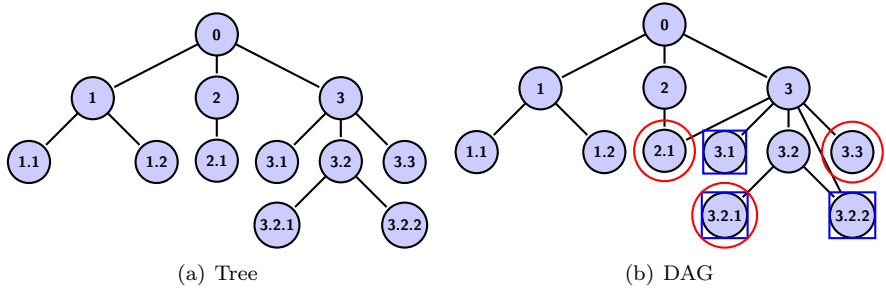

\end{LCA}

\noindent For example, in Figure \ref{Tree} $LCA(3.1, 3.2.2)$ = $3$. In the case of a DAG the definition of $LCA$ changes.
$LCA(n_{1},n_{2})$ is a set of nodes (instead of a single node), since it is possible for two nodes to have more than one $LCA$. 
Furthermore, the $LCA$ may not necessarily be the node that is furthest from the root. 
In order to define the LCA between two DAG nodes, we use the concept of the shortest path between them.

\newtheorem{path}[LCA]{Definition}
\begin{path}
Given a set $P_{all}(n_{1},n_{2})$ containing all paths that connect nodes $n_1$ and $n_2$, we define $paths_{min}(n_1,n_2) \subseteq P_{all}(n_{1},n_{2})$ as the set for which:\\
$\forall p \in paths_{min}(n_1,n_2); \nexists p' \in P_{all}(n_{1},n_{2}) \setminus paths_{min}(n_1,n_2) : cost(p) \leq cost (p')$

\end{path}

\noindent where the cost of a path corresponds to its length, when the edges of the hierarchy are unweighted.

For example in Figure \ref{DAG}: 
\begin{itemize}
 \item $path_{min}(2.1,3.1)=\{2.1, 3, 3.1\}$
 \item $path_{min}(2.1,3.2.2)=\{2.1, 3, 3.2.2\}$  
 \item $path_{min}(3.2.2, 3.2.1)=\{3.2.2, 3.2, 3.2.1\}$ 
\end{itemize}
It is worth noting that in the general case $paths_{min}(n_1,n_2)$ is a set of paths; not a single one.

In multi-label classification, we would like to extend the definition of $LCA$ to compare a node $n$ (e.g. a true class) 
against a set of nodes $S$ (e.g. the predicted classes). 

\newtheorem{LCAExp}[LCA]{Definition}
\begin{LCAExp}
The $LCA(n,S)$ of a node $n$ and a set of nodes $S$ is the set of all the lowest common ancestors $LCA(n,i)$ for each $i \in S_{best}(n,S) \subseteq S$, where
$S_{best}(n,S)= \{i \in S: \nexists j \in S,$ $j \neq i \land cost(path_{min}(n,i)) > cost(path_{min}(n,j))\}$
\end{LCAExp}
\noindent For example, in Figure \ref{DAG} $S_{best}(3.1,\{2.1, 3.3, 3.2.1\})=\{2.1, 3.3\}$ and \\
$LCA(3.1, \{2.1, 3.3, 3.2.1\})$ is $\{3\}$.

Given this definition and sets, $Y$ being the true and $\hat{Y}$ the predicted classes of an instance, we compute the 
$LCA (y, \hat{Y})$ of each element $y$ of $Y$. Similarly for each element $\hat{y}$ of $\hat{Y}$ by computing $LCA (\hat{y}, Y)$. Using Figure \ref{DAG}, let
$Y$ = \{2.1, 3.2.1, 3.3\} and $\hat{Y}$ = \{3.1, 3.2.1, 3.2.2\}. Then
\begin{itemize}
\item $LCA (2.1, \hat{Y}) = \{3\}$, connecting 2.1 with either 3.1 using $path_{min}(2.1,3.1)$ or 3.2.2 using $path_{min}(2.1,3.2.2)$.
\item $LCA (3.3, \hat{Y}) = \{3\}$, connecting 3.3 with either 3.1 using $path_{min}(3.3, 3.1)$ or 3.2.2 using $path_{min}(3.3, 3.2.2)$.
\item $LCA (3.2.1, \hat{Y}) = \{3.2.1\}$, connecting 3.2.1 with itself.
\item $LCA (3.2.1, Y) = \{3.2.1\}$, connecting 3.2.1 with itself.
\item $LCA (3.1, Y) = \{3\}$, connecting 3.1 with either 2.1 using $path_{min}(3.1, 2.1)$ or 3.3 using $path_{min}(3.1, 3.3)$.
\item $LCA (3.2.2, Y) = \{3.2, 3\}$, the first connecting 3.2.2 with 3.2.1 using \\
$path_{min}(3.2.2, 3.2.1)$ 
and the second connecting 3.2.2 with either
2.1 using $path_{min}(3.2.2, 2.1)$ or 3.3 using $path_{min}(3.2.2, 3.3)$.
\end{itemize}

Additionally, we are interested in the sets containing all the LCA of each of the two sets.
\newtheorem{LCAall}[LCA]{Definition}
\begin{LCAall}
Given a set of true classes (nodes) $Y$ and a set of predicted classes (nodes) $\hat{Y}$, we define $LCA_{all}(Y,\hat{Y})$ as the set containing all LCA($y,\hat{Y}$) 
for all $y \in Y$.
Similarly we define $LCA_{all}(\hat{Y},Y)$ as the set containing all LCA($\hat{y},Y$) for all $\hat{y} \in \hat{Y}$.
\end{LCAall}

\noindent In the above example $LCA_{all}(Y,\hat{Y})=$\{3, 3.2.1\}, $LCA_{all}(\hat{Y},Y)=$\{3, 3.2, 3.2.1\}.
\newtheorem{LCAGraph}[LCA]{Definition}
\begin{LCAGraph}
Given a set of true classes (nodes) $Y$, a set of predicted classes (nodes) $\hat{Y}$ and a set of $LCA_{all}$, we define $G^{ex}_{t}(Y,\hat{Y})$ as the graph that 
contains:
\begin{itemize}
 \item all $paths_{min}(y,a)$: $y \in Y \land a \in LCA(y,\hat{Y})$
 \item all $paths_{min}(y,a)$ subpaths of $paths_{min}(\hat{y}, y):\hat{y} \in \hat{Y} \land y \in S_{best}(\hat{y},Y) \land a \in LCA(\hat{y},Y)$ 
\end{itemize}
Similarly $G^{ex}_{p}(Y,\hat{Y})$ is the graph that contains:
\begin{itemize}
 \item all $paths_{min}(\hat{y},b)$: $\hat{y} \in \hat{Y}\land b \in LCA(\hat{y},Y)$
 \item all $paths_{min}(\hat{y},b)$ subpaths of $paths_{min}(y,\hat{y}):y \in Y \land \hat{y} \in S_{best}(y,\hat{Y}) \land b \in LCA(y,\hat{Y})$ 
\end{itemize}
\end {LCAGraph}

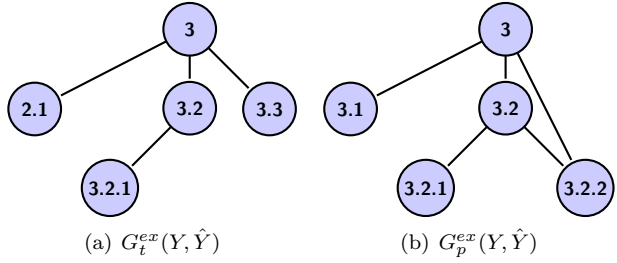
\begin{figure}[h]
\begin{center}
\subfigure[$G^{ex}_t(Y,\hat{Y})$] {\label{TDAGEX}
\begin{tikzpicture}[-,>=stealth',shorten >=1pt,auto,node distance=3cm, 
  thick,main node/.style={circle,fill=blue!20,draw,font=\sffamily\Large\bfseries, minimum size=14mm, scale=0.5}]

\node[main node] (1) {3};
\node[main node] (2) [below left of=1, xshift=-20mm] {2.1};
\node[main node] (3) [below of=1, yshift=9mm] {3.2};
\node[main node] (4) [below right of=1] {3.3};
\node[main node] (5) [below left of=3] {3.2.1};

\path[every node/.style={font=\sffamily\small}]
    (1) edge node [left] {} (2)	
	edge node [] {} (3)   
	edge node [right] {} (4)
    (3) edge node [left] {} (5);	
\end{tikzpicture}
}
\subfigure[$G^{ex}_p(Y,\hat{Y})$] {\label{PDAGEX}
\begin{tikzpicture}[-,>=stealth',shorten >=1pt,auto,node distance=3cm, 
  thick,main node/.style={circle,fill=blue!20,draw,font=\sffamily\Large\bfseries, minimum size=14mm, scale=0.5}]
\node[main node] (1) {3};
\node[main node] (2) [below left of=1, xshift=-20mm] {3.1};
\node[main node] (3) [below of=1, yshift=9mm] {3.2};
\node[main node] (4) [below right of=3] {3.2.2};
\node[main node] (5) [below left of=3] {3.2.1};

\path[every node/.style={font=\sffamily\small}]
    (1) edge node [left] {} (2)	
	edge node [] {} (3)   	
	edge node [right] {} (4)
    (3) edge node [right] {} (4)
        edge node [left] {} (5);	
    
\end{tikzpicture}
}
\caption{$G^{ex}_{t}(Y,\hat{Y})$ and $G^{ex}_{p}(Y,\hat{Y})$ augment the sets of the true and the predicted classes with $LCAs$.}
\label {DAGsEX}
\end{center}
\end{figure}

\noindent For example, for the $Y$ and $\hat{Y}$ of figure \ref{DAG} we get the $G^{ex}_{t}(Y,\hat{Y})$ and $G^{ex}_{p}(Y,\hat{Y})$ graphs of figure \ref{TDAGEX} and \ref{PDAGEX}, respectively.

Based on these graphs the true and predicted sets of classes are augmented, in order for the set-based measures to be calculated. 
In the case of Figure \ref{DAGsEX}, $Y_{aug}=\{3, 2.1, 3.2, 3.3, 3.2.1\}$ and 
$\hat{Y}_{aug}=\{3, 3.1, 3.2, 3.2.1, 3.2.2\}$. The next step is to calculate cost measures based on these two sets which in our case are the following:

\[
P_{LCA} = \frac{|\hat{Y}_{aug}\cap Y_{aug}|}{|\hat{Y}_{aug}|}
\]
\[
R_{LCA} = \frac{|\hat{Y}_{aug}\cap Y_{aug}|}{|Y_{aug}|}
\]
\[
F_{LCA} = \frac{2P_{LCA}R_{LCA}}{P_{LCA}+R_{LCA}}
\]

In the example of Figure \ref{DAGsEX} all three measures, $P_{LCA}$, $R_{LCA}$ and $F_{LCA}$, between sets $Y_{aug}$ and $\hat{Y}_{aug}$, are 0.6.
We prefer this approach over the symmetric difference loss, since it takes into account the TP in addition to FP and FN.
Ignoring TP leads systems to prefer predicting fewer categories, since missing a single FP usually costs more than the gain of finding an extra TP. 
This behavior is also observed in the results of real systems, (see section 4) and is considered undesirable.

\begin{figure}[h]
\begin{center}
\subfigure[$G_t(Y,\hat{Y})$] {\label{TDAG}
\begin{tikzpicture}[-,>=stealth',shorten >=1pt,auto,node distance=3cm, 
  thick,main node/.style={circle,fill=blue!20,draw,font=\sffamily\Large\bfseries, minimum size=14mm, scale=0.5}]

\node[main node] (1) {3};
\node[main node] (2) [below left of=1, xshift=-20mm] {2.1};
\node[main node] (3) [below right of=1] {3.3};
\node[main node] (4) [below right of=2] {3.2.1};

\path[every node/.style={font=\sffamily\small}]
    (1) edge node [left] {} (2)	
	edge node [right] {} (3);   
\end{tikzpicture}
}
\subfigure[$G_p(Y,\hat{Y})$] {\label{PDAG}
\begin{tikzpicture}[-,>=stealth',shorten >=1pt,auto,node distance=3cm, 
  thick,main node/.style={circle,fill=blue!20,draw,font=\sffamily\Large\bfseries, minimum size=14mm, scale=0.5}]

\node[main node] (1) {3};
\node[main node] (2) [below left of=1] {3.1};
\node[main node] (3) [below of=2] {3.2.1};
\node[main node] (4) [right of=3] {3.2.2};

\path[every node/.style={font=\sffamily\small}]
    (1) edge node [left] {} (2)
	edge node [right] {} (4);
    
\end{tikzpicture}
}
\caption{$G_{t}(Y,\hat{Y})$ and $G_{p}(Y,\hat{Y})$ are the minimal graphs augmenting the sets of the true and the predicted classes with $LCAs$.}
\label {DAGs}
\end{center}
\end{figure}
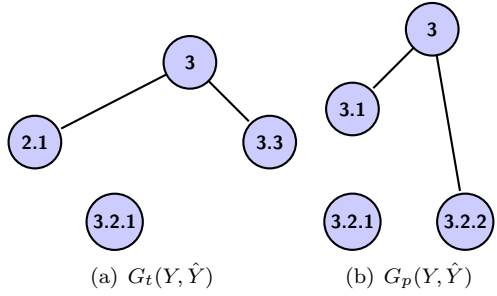

The two graphs $G^{ex}_{t}(Y,\hat{Y})$ and $G^{ex}_{p}(Y,\hat{Y})$ were created using all nodes of $LCA_{all}(Y,\hat{Y})$ and $LCA_{all}(\hat{Y},Y)$ and all corresponding paths. 
However, subgraphs of the two graphs $G_{t}(Y,\hat{Y})\subseteq G^{ex}_{t}(Y,\hat{Y})$ and $G_{p}(Y,\hat{Y}) \subseteq G^{ex}_{p}(Y,\hat{Y})$, could be selected that would connect 
each node of $Y \cup \hat{Y}$ with an $LCA$
and  vice versa. For example, in Figure \ref{DAG} node 3.2.2 has two LCAs, node 3.2 and 3. Node 3.2 could be removed from $G^{ex}_{t}(Y,\hat{Y})$ and $G^{ex}_{p}(Y,\hat{Y})$, without 
breaking the condition of any 
node in $Y \cup \hat{Y}$ with an LCA or vice versa. We would then get graphs
$G_{t}(Y,\hat{Y})$ and $G_{p}(Y,\hat{Y})$ of Figure \ref{DAGs}. $P_{LCA}$, $R_{LCA}$ and $F_{LCA}$, between the reduced sets $Y_{aug}$ and $\hat{Y}_{aug}$ of Figure \ref{DAGs}, are 0.5 instead of 0.6 (Figure \ref{DAGsEX}).

In other words graphs $G_{t}(Y,\hat{Y})$ and $G_{p}(Y,\hat{Y})$ should comprise the nodes necessary for connecting the nodes of the two sets, through their LCAs. 
Redundant nodes can lead to fluctuations in $P_{LCA}$, $R_{LCA}$ and $F_{LCA}$, 
and should be removed. In order to obtain the minimal LCA graphs, we have to solve the following maximization problem:

\begin{myproblem}{Minimal LCA graph extension.}
\begin{eqnarray*}
\left\{
\begin{array}{ll}
 \displaystyle\argmax_{\substack { {(G_{t}(Y,\hat{Y})\subseteq G^{ex}_{t}(Y,\hat{Y})}, \\
{G_{p}(Y,\hat{Y}) \subseteq G^{ex}_{p}(Y,\hat{Y})})}} 
{F_{LCA}(G_{t}(Y,\hat{Y}),G_{p}(Y,\hat{Y}))}\\ 
 \mbox{subject to:} & \\
 
 \hspace{0.5cm} \mbox{(i)} \,\,\, \forall y \in Y; y \in G_{t}(Y,\hat{Y})  \mbox{ and } \forall \hat{y} \in \hat{Y}; \hat{y} \in G_{p}(Y,\hat{Y}) & \\
 \hspace{0.5cm} \mbox{(ii)} \,\,\, \forall y \in Y; \exists a: a\in G_{t}(Y,\hat{Y}) \land a\in G_{p}(Y,\hat{Y}) \land a \in LCA(y,\hat{Y}) \\
 \hspace{1cm} \,\,\, \forall \hat{y} \in \hat{Y}; \exists b: b\in G_{t}(Y,\hat{Y}) \land b\in G_{p}(Y,\hat{Y}) \land b \in LCA(\hat{y},{Y}) \\
 \hspace{0.5cm} \mbox{(iii)} \,\,\, \not \exists G'_{t}(Y,\hat{Y}) \mbox{ subject to constraints (i) and (ii)} :\\
 \hspace{3.3cm} |\{a: a\in G_{t}(Y,\hat{Y}) \land a \in LCA_{all}(Y,\hat{Y}) \}| > \\ 
 \hspace{3.3cm} |\{a':a' \in G'_{t}(Y,\hat{Y}) \land a' \in LCA_{all}(\hat{Y},Y)\}|\\
 \hspace{1.2cm} \,\,\, \not \exists G'_{p}(Y,\hat{Y})\mbox{ subject to constraints (i) and (ii)} : \\
 \hspace{3.3cm} |\{b:b \in G_{p}(Y,\hat{Y}) \land b \in LCA_{all}(Y,\hat{Y}) \}| >\\
 \hspace{3.3cm} |\{b':b' \in G'_{p}(Y,\hat{Y}) \land b' \in LCA_{all}(\hat{Y},Y)\}|\\  
 \hspace{0.5cm} \mbox{(iv)} \,\,\, \forall y \in Y; \exists p_{y} \in G_{t}(Y,\hat{Y}): p_{y} \in paths_{min}(y,a) \land a \in LCA_{all}(Y,\hat{Y}) \\
 \hspace{1.1cm} \,\,\, \forall \hat{y} \in \hat{Y}; \exists p_{\hat{y}} \in G_{p}(Y,\hat{Y}): p_{\hat{y}} \in paths_{min}(\hat{y},b) \land b \in LCA_{all}(\hat{Y},Y) \\
 \hspace{0.5cm} \mbox{(v)} \,\,\, \forall a \in G_{t}(Y,\hat{Y}) \cap LCA_{all}(\hat{Y},Y); \exists p_{y} \in G_{t}(Y,\hat{Y}):\\
 \hspace{3.3cm} p_{y} \in paths_{min}(y,a) \land y \in Y \\
 \hspace{0.95cm} \,\,\, \forall b \in G_{p}(Y,\hat{Y}) \cap LCA_{all}(Y,\hat{Y}); \exists p_{\hat{y}}  \in G_{p}(Y,\hat{Y}): \\
 \hspace{3.3cm} p_{\hat{y}}  \in path_{min}(\hat{y},b) \land \hat{y} \in \hat{Y}
 \end{array}
\right.
\end{eqnarray*}
\end{myproblem}   

The maximization of $F_{LCA}(G_{t}(Y,\hat{Y}),G_{p}(Y,\hat{Y}))$ is subject to a set of constraints:
Constraint (i) requires all class nodes of an initial set ($Y$ or $\hat{Y}$) to be included in the final subgraphs.
Constraint (ii) enforces the existence of at least one LCA for each node of $Y \cup \hat{Y}$, in the subgraphs. 
Constraint (iii) limits the total number of LCAs used to the minimum required in order to be able to satisfy constraints (i) and (ii). 
Constraint (iv) implies the existence of at least one path connecting each class node of each subgraph to one of its LCAs, 
while constraint (v) implies the inverse, i.e. that each LCA of the subgraphs is connected with at least one class node of each subgraph.

One way to solve this maximization problem would be to create all possible $G_{t}(Y,\hat{Y})$ and $G_{p}(Y,\hat{Y})$ graphs, which respect the above constraints and 
choose the ones leading to the highest $F_{LCA}$. This procedure is very computationally expensive and for this reason we devised the approximation 
presented in algorithm \ref{appro}. 
  
\begin{algorithm} [H]
\caption {Approximation algorithm for computing $G_{t}(Y,\hat{Y})$ and $G_{p}(Y,\hat{Y})$, subgraphs of $G^{ex}_{t}(Y,\hat{Y})$ and $G^{ex}_{p}(Y,\hat{Y})$}
\label{appro}
\begin{algorithmic}[1]
\Procedure{GetSubgraphs}{$G^{ex}_{t}(Y,\hat{Y})$, $G^{ex}_{p}(Y,\hat{Y})$}
\State $LCA_{all} \gets getLCAsFrom(G^{ex}_{t}(Y,\hat{Y}),G^{ex}_{p}(Y,\hat{Y}))$
\State $bestLCAs \gets \Call{GetBestLCAs}{LCA_{all},G^{ex}_{t}(Y,\hat{Y}),G^{ex}_{p}(Y,\hat{Y})$}
\State $finalPaths \gets \Call{GetBestPaths}{bestLCAs,G^{ex}_{t}(Y,\hat{Y}),G^{ex}_{p}(Y,\hat{Y})$}
\State $G_{t}(Y,\hat{Y}) \gets$ all paths from $finalPaths$ containing a node $\in Y$
\State $G_{p}(Y,\hat{Y}) \gets$ all paths from $finalPaths$ containing a node $\in \hat{Y}$
\EndProcedure
\Procedure{GetBestLCAs}{$LCAs$, $G_t$, $G_p$}
\State $sortedLCAs \gets sort(LCAs)$\Comment{sort LCAs, by the number of $Y$ and $\hat{Y}$ that they connect, in descending order}
\State $bestLCAs \gets \{\}$, $i\gets 1$
\Repeat
  \State $bestLCAs\gets bestLCAs \cup sortedLCAs_{i}$
  \State $i\gets i + 1$
\Until {\Call{Satisfied}{$bestLCAs$, $G_t(Y,\hat{Y})$, $G_p(Y,\hat{Y})$}}
\Comment {procedure SATISFIED checks whether constraints (i), (ii) and (iii) of the optimization problem are satisfied}
\For{$i\gets 1$ to $sizeof(bestLCAs)$} \Comment {top-down redundancy removal}
  \If {\Call{Satisfied}{$bestLCAs \setminus bestLCAs_{i}$, $G_t(Y,\hat{Y})$, $G_p(Y,\hat{Y})$}}
  \State $bestLCAs\gets bestLCAs \setminus bestLCAs_{i}$
  \EndIf
\EndFor

\For{$i\gets sizeof(bestLCAs)$ to $1$} \Comment {bottom-up redundancy removal}
  \If {\Call{Satisfied}{$bestLCAs \setminus bestLCAs_{i}$, $G_t(Y,\hat{Y})$, $G_p(Y,\hat{Y})$}}
  \State $bestLCAs\gets bestLCAs \setminus bestLCAs_{i}$
  \EndIf
\EndFor
\State \textbf{return $bestLCAs$}
\EndProcedure
\algstore{bkbreak}
\end{algorithmic}
\end{algorithm}

\begin{algorithm}[h]
\caption{Part 2 of Algorithm \ref{appro}}
\begin{algorithmic}[1]
\algrestore{bkbreak}
\Procedure{GetBestPaths}{$bestLCAs$, $G_t(Y,\hat{Y})$, $G_p(Y,\hat{Y})$}
\State $bestY \gets \{\}$, $best\hat{Y} \gets \{\}$
\For {$i\gets 1$ to $sizeof(bestLCAs)$}  
  \For {$j\gets 1,sizeof(Y)$} \Comment {find nodes $Y_j$, for which $bestLCA_i$ is an LCA} 
  \If {$bestLCAs_{i} \in LCAs(Y_{j},\hat{Y})$}
    \State $bestY \gets bestY  \cap \Call{bestPath}{bestLCAs_{i},Y_{j}}$ 
    \Comment {BestPath selects the path of $path_{min}(bestLCA_{i},Y_{j})$ that shares most common nodes with all other selected paths}
  \EndIf
  \EndFor
  \For {$j\gets 1$ to $sizeof(\hat{Y})$} 
  \If {$bestLCAs_{i} \in LCAs(\hat{Y_{j}},Y)$}
    \State $best\hat{Y} \gets best\hat{Y} \cap \Call{BestPath}{bestLCAs_{i},\hat{Y_{j}}}$ 
  \EndIf
  \EndFor
\EndFor
\State \textbf{return $bestY \cup best\hat{Y}$}
\EndProcedure
\end{algorithmic}
\end{algorithm}

The main procedure is decomposed into three subprocedures. Procedure GetBestLCAs returns an approximation of the minimum amount of $LCAs$ needed in order 
to satisfy constraints (i), (ii) and (iii) of the maximization problem.
This is achieved by initially sorting, in descending order, all $LCAs$ by the number of $Y$ and $\hat{Y}$ that they connect. On this list, we perform two passes, first top-down  
and then bottom-up removing all redundant $LCAs$, i.e $LCAs$ of nodes for which other $LCAs$ are already included in the list. In the final step of the algorithm, GetBestPaths selects the minimum 
paths that satisfy constraints (iv) and (v). In case two or more paths exist that connect the same node with an LCA we choose the one which leads to the smallest
possible subgraphs.

An interesting issue arises when a class and one of its ancestors co-exist in the predicted or the true class sets. Assume for example a system A predicting that an instance 
belongs to node $X$, while another system B assigns it also to one of the ancestors of $X$. Each extra ancestor
of $X$ would lead to higher $F_{1}$ score since it would increase the size of the $Y_{aug} \cap \hat{Y}_{aug}$ set. This happens because all the ancestors of an $LCA(n_{1},x_{2})$ are 
also ancestors of nodes $n_{1}$ and $n_{2}$. We address this issue by removing from set $Y$ 
any node $y$ for which $\exists y'\in Y : y'$ is a descendant of $y$. We then do the same for set $\hat{Y}$ by removing each node $\hat{y}$ 
for which $\exists \hat{y'} \in \hat{Y} : \hat{y'}$ is a descendant of $\hat{y}$.

LCA precision, recall and $F_1$ are not purely set-based measures, but are actually a bridge between pair-based and set-based measures. They are based on  
augmented sets of predicted and true nodes and calulate scores, based on the relation of those two sets. However the use of the LCA(n,S) 
leads to a pairing of predicted and true nodes. So these measures could also be characterized as a hybrid, 
combining the advantages of both types of measure.

\section{Case Studies}

In this section we apply various measures to selected cases in order to demonstrate their pros and cons. As a representative of the pair-based measures we 
chose the Graph Induced Error (GIE), while for set-based ones we selected the hierarchical versions of precision ($P_H$), recall ($R_H$), $F_1$-measure ($F_H=\tfrac{2 \cdot P_H \cdot R_H}{P_H+R_H}$) and 
Symmetric Difference Loss ($l_{\Delta}(Y_{aug},\hat{Y}_{aug})$), 
using all the ancestors of the predicted ($\hat{Y}$) and true ($Y$) labels in order to augment the sets of classes. We also use our proposed pair-based measure MGIA and the set-based 
LCA versions of precision ($P_{LCA}$), recall ($R_{LCA}$), $F_1$-measure ($F_{LCA}$) in order to illustrate their advantages 
and limitations, as well as the differences between the two types of measure. Regarding MGIA we also provide in parenthesis the $fnerror$, before the transformation that we propose in subsection 2.3.3, 
in order to be easier comparable with GIE. All the above measures are implemented in a fast and easy to use tool written in 
C++ that is open source and available for download.\footnote{The tool is available from http://nlp.cs.aueb.gr/software\_and\_datasets/HEMKit.zip} 

Like all pair-based methods GIE and MGIA require a maximum distance threshold, above which nodes are paired with a default one. 
In the cases that we study here, this threshold is set to 5.

Based on the situations presented in section 2.2, which highlight important challenges in hierarchical evaluation, the case studies here correspond to 
specific examples where these situations appear. The list of cases here is not exhaustive, but it is sufficient to motivate the use of the proposed measures. 
Additionally, in the Section 4 we present results on real datasets and classification systems.

\subsection{Handling the Pairing Problem}

\begin{figure}[ht]
\begin{center} 
\subfigure[] {\label{case_1a}
\begin{tikzpicture}[-,>=stealth',shorten >=1pt,auto,node distance=3cm,
  thick,main node/.style={circle,fill=blue!20,draw,font=\sffamily\Large\bfseries, minimum size=11mm, scale=0.7}]

\node[main node] (1) {A};
\node[main node] (2) [below left of=1] {B};
\node[main node] (3) [below right of=1] {C};
\node[main node] (4) [below left of=2] {$T_1$};
\node[main node] (5) [below of=2,yshift=9mm] {$P_1$};
\node[main node] (6) [below right of=2] {$P_2$};

\path[every node/.style={font=\sffamily\small}]
    (1) edge node [left] {} (2)
	edge node [right] {} (3)
    (2) edge node [left] {} (4)
	edge node [] {} (5)
	edge node [right] {} (6);
\end{tikzpicture}
}
\subfigure[] {\label{case_1b}
\begin{tikzpicture}[-,>=stealth',shorten >=1pt,auto,node distance=3cm,
  thick,main node/.style={circle,fill=blue!20,draw,font=\sffamily\Large\bfseries, minimum size=11mm, scale=0.7}]

\node[main node] (1) {A};
\node[main node] (2) [below left of=1] {B};
\node[main node] (3) [below right of=1] {C};
\node[main node] (4) [below left of=2] {$T_1$};
\node[main node] (5) [below of=2,yshift=9mm] {$T_2$};
\node[main node] (6) [below right of=2] {$P_1$};

\path[every node/.style={font=\sffamily\small}]
    (1) edge node [left] {} (2)
	edge node [right] {} (3)
    (2) edge node [left] {} (4)
	edge node [] {} (5)
	edge node [right] {} (6);
\end{tikzpicture}
}
\caption{Different numbers of true and predicted labels}
\label{case1}
\end{center}
\end{figure}
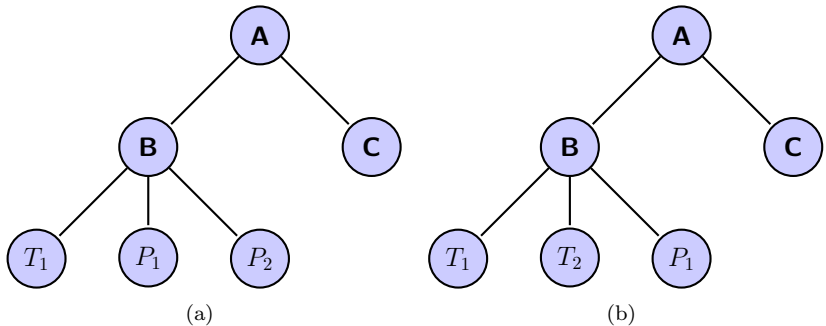

Case 1 captures the situation where the number of true $T$ and predicted $P$ labels (classes) differ, while being at the same level of the hierarchy. In this elementary case, Figures \ref{case_1a} 
and \ref{case_1b} are two symmetric variants leading to different, but symmetric hierarchical precision ($P_H$) and recall ($R_H$) scores, as shown in Table \ref{case1t}. The same is true for 
our proposed LCA versions of precision and recall ($P_{LCA}$ and $R_{LCA}$). The results between the hierarchical version and the LCA versions differ because the LCA versions ignore the graph above node B
which is the lowest common ancestor of nodes $T1$, $P_1$ and $P_2$. The hierarchical versions on the other hand also take into account node A and in that way they give higher results. 
This behavior is undesirable, since each node above B would increase the results of the hierarchical measures but would not affect the LCA versions.

All other measures give the same result in both cases. $l_{\Delta}(Y_{aug},\hat{Y}_{aug})$ always computes the symmetric difference between the two augmented sets. The symmetric difference takes into 
account only the sum of false possitives and false negatives which are the same in both cases. Among the pair-based measures, GIE seems inappropriate for this problem. It matches $T_1$ with one of $P_1$ and $P_2$ in Fig \ref{case_1a} and penalizes 
the unmatched predicted class with the maximum cost, ignoring the fact that the misclasification is in the proximity of the correct category. Similarly 
in Figure \ref{case_1b} MGIA provides a more suitable evaluation, as it allows multiple categories to match to the nearest one.

\begin{table} [ht]
\begin{center}
\fitMe{
\begin{tabular}{|c c c c c c c c c c|}
\hline
& GIE & MGIA & $P_H$ & $R_H$ & $F_H$ & $l_{\Delta}(Y_{aug},\hat{Y}_{aug})$ & $P_{LCA}$ & $R_{LCA}$ & $F_{LCA}$\\
\hline
a & 7 & 0.73(4) & 0.5 & 0.66 & 0.57 & 3 & 0.33 & 0.5 & 0.4\\
b & 7 & 0.73(4) & 0.66 & 0.5 & 0.57 & 3 & 0.5 & 0,33 & 0.4\\
\hline
\end{tabular}
}
\end{center}
\caption{Results per measure for Figure \ref{case1}}
\label{case1t}
\end{table}

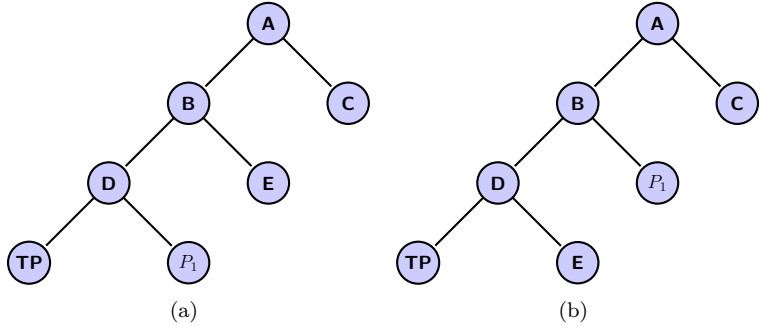
\begin{figure}[ht]
\begin{center}
\subfigure[] {\label{case_2a}
\begin{tikzpicture}[-,>=stealth',shorten >=1pt,auto,node distance=3cm,
  thick,main node/.style={circle,fill=blue!20,draw,font=\sffamily\Large\bfseries, minimum size=11mm, scale=0.5}]

\node[main node] (1) {A};
\node[main node] (2) [below left of=1] {B};
\node[main node] (3) [below right of=1] {C};
\node[main node] (4) [below left of=2] {D};
\node[main node] (5) [below right of=2] {E};
\node[main node] (6) [below right of=4] {$P_1$};
\node[main node] (7) [below left of=4] {TP};

\path[every node/.style={font=\sffamily\small}]
    (1) edge node [left] {} (2)
	edge node [right] {} (3)
    (2) edge node [left] {} (4)	
	edge node [right] {} (5)
    (4) edge node [left] {} (6)	
	edge node [right] {} (7);    		
\end{tikzpicture}
}
\subfigure[] {\label{case_2b}
\begin{tikzpicture}[-,>=stealth',shorten >=1pt,auto,node distance=3cm,
  thick,main node/.style={circle,fill=blue!20,draw,font=\sffamily\Large\bfseries, minimum size=11mm, scale=0.5}]

\node[main node] (1) {A};
\node[main node] (2) [below left of=1] {B};
\node[main node] (3) [below right of=1] {C};
\node[main node] (4) [below left of=2] {D};
\node[main node] (5) [below right of=2] {$P_1$};
\node[main node] (6) [below right of=4] {E};
\node[main node] (7) [below left of=4] {TP};

\path[every node/.style={font=\sffamily\small}]
    (1) edge node [left] {} (2)
	edge node [right] {} (3)
    (2) edge node [left] {} (4)	
	edge node [right] {} (5)
    (4) edge node [left] {} (6)	
	edge node [right] {} (7);    		
\end{tikzpicture}
}
\caption{Different numbers of true and predicted labels. Distance from closest true class differs}
\label{case2}
\end{center}
\end{figure}

Case 2 in Figure \ref{case2} is an example showing that taking into account all the ancestors is undesirable compared to our proposed LCA approach for set based measures. The hierarchy is still a tree 
and the classification is multi-label. $TP$ is a node which was predicted correctly, while $P_1$ is misclassified. Although the mistake in Figure \ref{case_2b} is worse than that of 
Figure \ref{case_2a}, since it is further from the true class $TP $, all set-based measures except our proposed LCA measures continue to give the same results. This is because $l_{\Delta}(Y_{aug},\hat{Y}_{aug})$ 
and the hierarchical versions of precision, recall and $F_1$ take into account all the ancestors of the predicted and true labels, while the LCA versions, which are hybrid measures,
uses the augmented graphs $G_t(Y,\hat{Y})$ and $G_p(Y,\hat{Y})$ which were created using only the least common ancestors (LCAs). LCA measures do some kind of
pairing in order to pair each node with the closest node of the other set and in that way take into account the distance between predicted and true nodes, which the other set-based measure 
ignore. 

\begin{table} [ht]
\begin{center}
\fitMe{
\begin{tabular}{|c|c c c c c c c c c|}
\hline
& GIE & MGIA & $P_H$ & $R_H$ & $F_H$ & $l_{\Delta}(Y_{aug},\hat{Y}_{aug})$ & $P_{LCA}$ & $R_{LCA}$ & $F_{LCA}$\\
\hline
a & 2 & 0.8(2) & 0.8 & 1 & 0.89 & 1 & 0.66 & 1 & 0.8\\
b & 3 & 0.7(3) & 0.8 & 1 & 0.89 & 1 & 0.66 & 0.66 & 0.66\\
\hline
\end{tabular}
}
\end{center}
\caption{Results per measure for Figure \ref{case2}}
\label{case2t}
\end{table}

Thus in Figure \ref{case_2a} the augmented sets of the LCA are \{TP, D\} and \{TP, D, $P_1$\} while for all other set-based measures are \{TP, D, B, A\} and \{TP, $P_1$, D, B, A\}. In Figure 
\ref{case_2b} the augmented sets of LCA become \{TP, D, B\} and \{TP, B, $P_1$\} while for all other set-based measures they remain the same.
The differentiation between such cases is an advantage of LCA-SDL over existing set-based measures.

\subsection{Handling alternative paths}

\begin{figure}[ht]
\begin{center}
\subfigure[] {\label{case_3a}
\begin{tikzpicture}[-,>=stealth',shorten >=1pt,auto,node distance=3cm,
  thick,main node/.style={circle,fill=blue!20,draw,font=\sffamily\Large\bfseries, minimum size=11mm, scale=0.5}]

\node[main node] (1) {A};
\node[main node] (2) [below left of=1] {B};
\node[main node] (3) [below right of=1] {C};
\node[main node] (4) [below left of=2] {$T_1$};
\node[main node] (5) [below right of=2] {$P_1$};

\path[every node/.style={font=\sffamily\small}]
    (1) edge node [left] {} (2)
	edge node [right] {} (3)
    (2) edge node [left] {} (4)
	edge node [right] {} (5)
    (3) edge node [left] {} (5);	
\end{tikzpicture}
}
\subfigure[] {\label{case_3b}
\begin{tikzpicture}[-,>=stealth',shorten >=1pt,auto,node distance=3cm,
  thick,main node/.style={circle,fill=blue!20,draw,font=\sffamily\Large\bfseries, minimum size=11mm, scale=0.5}]

\node[main node] (1) {A};
\node[main node] (2) [below left of=1] {B};
\node[main node] (3) [below left of=2] {$T_1$};
\node[main node] (4) [below right of=2] {$P_1$};

\path[every node/.style={font=\sffamily\small}]
    (1) edge node [left] {} (2)	
    (2) edge node [left] {} (3)
	edge node [right] {} (4);    	
\end{tikzpicture}
}
\caption{Many paths of same length leading A to the same node $P_1$}
\label {case3}
\end{center}
\end{figure}
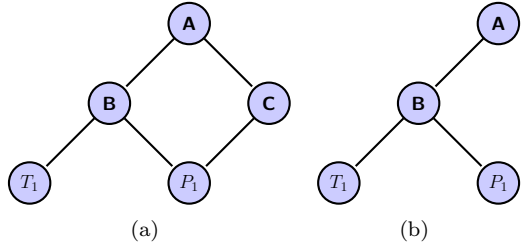

In Figure \ref{case_3a} we assume a single-label classification task, where the hierarchy is a $DAG$. In this $DAG$ there are two paths from the root (node $A$) to node $P_1$. It is worth noting 
that this is the simplest case, where both paths \{A, B, $P_1$\} and \{A, C, $P_1$\} have the same length. In this case, pair-based measures (Table \ref{case3t}) remain
unaffected compared to Figure \ref{case_3b} which is a tree, a desirable behavior that is due to the use of the shortest path from $T_1$ to $P_1$, while existing set-based measures are affected. 
In particular, they calculate a misclassification error that takes into account both of the two alternative paths to $P_1$. 
On the other hand LCA measures behave like the pair-based measures, due to the use of the lowest common ancestor, having again an advantage over existing set-based measures. 

\begin{table} [ht]
\begin{center}
\fitMe{
\begin{tabular}{| c c c c c c c c c c|}
\hline
& GIE & MGIA & $P_H$ & $R_H$ & $F_H$ & $l_{\Delta}(Y_{aug},\hat{Y}_{aug})$ & $P_{LCA}$ & $R_{LCA}$ & $F_{LCA}$\\
\hline
a & 2 & 0.8(2) & 0.5 & 0.66 & 0.57 & 3 & 0.5 & 0.5 & 0.5\\
b & 2 & 0.8(2) & 0.66 & 0.66 & 0.66 & 2 & 0.5 & 0.5 & 0.5\\
\hline
\end{tabular}
}
\end{center}
\caption{Results per measure for Figure \ref{case3}}
\label{case3t}
\end{table}

\subsection{Combining elementary cases}

\begin{figure}[h]
\begin{center}
\begin{tikzpicture}[-,>=stealth',shorten >=1pt,auto,node distance=3cm,
  thick,main node/.style={circle,fill=blue!20,draw,font=\sffamily\Large\bfseries, minimum size=11mm, scale=0.5}]

\node[main node] (1) {A};
\node[main node] (2) [below left of=1] {B};
\node[main node] (3) [below right of=1] {C};
\node[main node] (4) [below left of=2] {$T_1$};
\node[main node] (5) [below right of=2] {$P_1$};
\node[main node] (6) [below of=3, yshift=9mm] {D};
\node[main node] (7) [below right of=3] {$P_2$};

\path[every node/.style={font=\sffamily\small}]
    (1) edge node [left] {} (2)
	edge node [right] {} (3)
    (2) edge node [left] {} (4)
	edge node [right] {} (5)
    (3) edge node [left] {} (5)
	edge node [] {} (6)
	edge node [right] {} (7);
\end{tikzpicture}
\caption{Combining the pairing problem with alternative paths in a single example, where none of the set-based methods is affected.}
\label{case4}
\end{center}
\end{figure}
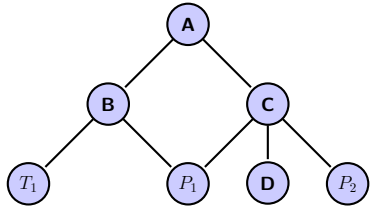

In this subsection we study cases where the combination of the above mentioned problems leads to variable behavior of the set-based evaluation measures. 
Figure \ref{case4} presents such a case, where although GIE is affected by the matching problem discussed in case 1, the set-based methods give 
similar results to each other according to Table \ref{case4t}.
The multiple paths phenomenon does not affect the hierarchical versions of precision, recall and $F_1$ because all nodes above the lowest common 
ancestors $B$ and $A$ of $T_1$, $P_1$ and $P_2$ are also shared by all classes (true and predicted).
With this example we wish to show that there are certain cases in which existing set-based methods give the same results as the LCA ones, but 
we have not identified cases in which the behavior of an existing set-based measure would be more desirable than that of the LCA versions.

\begin{table} [h]
\begin{center}
\fitMe{
\begin{tabular}{|c c c c c c c c c|}
\hline
GIE & MGIA & $P_H$& $R_H$ & $F_H$ & $l_{\Delta}(Y_{aug},\hat{Y}_{aug})$ & $P_{LCA}$ & $R_{LCA}$ & $F_{LCA}$\\
\hline
7 & 0.6(6) & 0.4 & 0.66 & 0.5 & 4 & 0.4 & 0.66 & 0.5\\
\hline
\end{tabular}
}
\end{center}
\caption{Results per measure for Figure \ref{case4}}
\label{case4t}
\end{table}

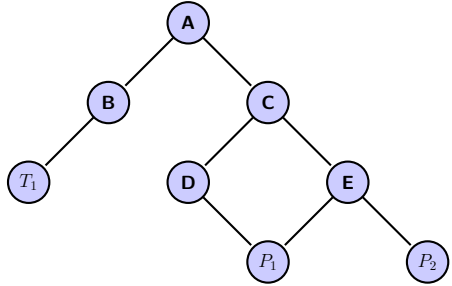
\begin{figure}[ht]
\begin{center}
\begin{tikzpicture}[-,>=stealth',shorten >=1pt,auto,node distance=3cm,
  thick,main node/.style={circle,fill=blue!20,draw,font=\sffamily\Large\bfseries, minimum size=11mm, scale=0.5}]

\node[main node] (1) {A};
\node[main node] (2) [below left of=1] {B};
\node[main node] (3) [below right of=1] {C};
\node[main node] (4) [below left of=2] {$T_1$};
\node[main node] (5) [below left of=3] {D};
\node[main node] (6) [below right of=3] {E};
\node[main node] (7) [below left of=6] {$P_1$};
\node[main node] (8) [below right of=6] {$P_2$};

\path[every node/.style={font=\sffamily\small}]
    (1) edge node [left] {} (2)
	edge node [right] {} (3)
    (2) edge node [left] {} (4)	
    (3) edge node [left] {} (5)
	edge node [right] {} (6)
    (5) edge node [right] {} (7)
    (6) edge node [left] {} (7)
	edge node [right] {} (8);
\end{tikzpicture}
\caption{Combining the pairing problem with alternative paths in a single example where all previous set-based measures behave undesirably, while our proposed LCA measures do not.}
\label{case6}
\end{center}
\end{figure}

The case shown in Figure \ref{case6} differs from the previous case, since the right sub-graph now connects to the left one only through node $A$ and also $P_1$ connects to $C$ through two different 
nodes $D$ and $E$. This is the reason why now hierarchical versions of precision, recall and $F_1$ differ from the LCA versions, as shown in table \ref{case6t}. LCA versions, due to its 
nearest common ancestor approach, ignores node $D$ while all other set-based measures count both D and E and thus over-penalize the error of $P_1$.  

\begin{table} [ht]
\begin{center}
\fitMe{
\begin{tabular}{|c c c c c c c c c|}
\hline
GIE & MGIA & $P_H$ & $R_H$ & $F_H$ & $l_{\Delta}(Y_{aug},\hat{Y}_{aug})$ & $P_{LCA}$ & $R_{LCA}$ & $F_{LCA}$\\
\hline
10 & 0(10) & 0.166 & 0.33 & 0.22 & 7 & 0.2 & 0.33 & 0.25\\
\hline
\end{tabular}
}
\end{center}
\caption{Results per measure for Figure \ref{case6}}
\label{case6t}
\end{table}

\subsection{Multiple path counting}

\begin{figure}[h]

\subfigure[] {\label{case_7a}
\begin{tikzpicture}[-,>=stealth',shorten >=1pt,auto,node distance=3cm,
  thick,main node/.style={circle,fill=blue!20,draw,font=\sffamily\Large\bfseries, minimum size=11mm, scale=0.5}]

\node[main node] (1) {A};
\node[main node] (2) [below left of=1] {B};
\node[main node] (3) [below right of=1] {C};
\node[main node] (4) [below left of=2] {$T_1$};
\node[main node] (5) [below right of=2] {$P_1$};
\node[main node] (6) [below right of=3] {D};
\node[main node] (7) [below left of=6] {E};
\node[main node] (8) [below right of=6] {$P_2$};

\path[every node/.style={font=\sffamily\small}]
    (1) edge node [left] {} (2)
	edge node [right] {} (3)
    (2) edge node [left] {} (4)	
	edge node [right] {} (5)
    (3) edge node [right] {} (6)
    (6) edge node [left] {} (7)
	edge node [right] {} (8);  
\end{tikzpicture}
}
\subfigure[] {\label{case_7b}
\begin{tikzpicture}[-,>=stealth',shorten >=1pt,auto,node distance=3cm,
  thick,main node/.style={circle,fill=blue!20,draw,font=\sffamily\Large\bfseries, minimum size=11mm, scale=0.48}]

\node[main node] (1) {A};
\node[main node] (2) [below left of=1] {B};
\node[main node] (3) [below right of=1] {C};
\node[main node] (4) [below left of=2] {$T_1$};
\node[main node] (5) [below right of=2] {E};
\node[main node] (6) [below right of=3] {D};
\node[main node] (7) [below left of=6] {$P_1$};
\node[main node] (8) [below right of=6] {$P_2$};

\path[every node/.style={font=\sffamily\small}]
    (1) edge node [left] {} (2)
	edge node [right] {} (3)
    (2) edge node [left] {} (4)	
	edge node [right] {} (5)
    (3) edge node [right] {} (6)
    (6) edge node [left] {} (7)
	edge node [right] {} (8);  
\end{tikzpicture}
}
\caption{Comparison between pair-based and set-based measures, counting paths more than once.}
\label {case7}
\end{figure}
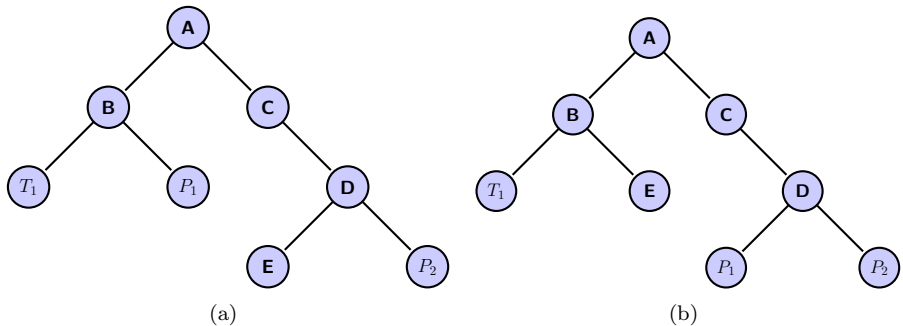

Due to comparison of pairs of true and predicted classes, pair-based methods will often count the same path more than once. This multiple counting increases 
the error estimated by these methods. Figure \ref{case7}, illustrates such a case. In Figure \ref{case_7a} the edge between $T_1$ and $B$ 
will be counted for the length of both \{$T_1$, B, $P_1$\} and \{$T_1$, B, A, C, D, $P_2$\} paths. 
As a result, pair-based measures in this case tend to overestimate the errors, in comparison to set-based ones. For each extra predicted node that we add as a descendant of $B$ 
pair-based error estimates increase by at least 2 while the size of $Y_{aug}$ of set-based measures increase by 1 which seems more reasonable for such a change in the hierarchy.

\begin{table} [h]
\begin{center}
\fitMe{
\begin{tabular}{|c | c c c c c c c c c|}
\hline
& GIE & MGIA & $P_H$ & $R_H$ & $F_H$ & $l_{\Delta}(Y_{aug},\hat{Y}_{aug})$ & $P_{LCA}$ & $R_{LCA}$ & $F_{LCA}$\\
\hline
a & 7 & 0.533(7) & 0.33 & 0.66 & 0.44 & 5 & 0.33 & 0.66 & 0.44 \\
b & 10 & 0.33(10) & 0.2 & 0.33 & 0.25 & 6 & 0.2 & 0.33 & 0.25 \\
\hline
\end{tabular}
}
\end{center}
\caption{Results per measure for Figure \ref{case7}}
\label{case7t}
\end{table}

Figure \ref{case_7b} presents a similar example. According to Table \ref{case7t}, the error of MGIA before the proposed transformation is increased from 7 to 10, while $l_{\Delta}(Y_{aug},\hat{Y}_{aug})$ increases 
from 5 to 6 and $F_{LCA}$ decreases from 0.44 to 0.25. 
This is because the whole path form $T_1$ to $D$ is counted twice by the pair-based method. However double counting seems desirable in this case, as the error in \ref{case_7b} is more severe than that in \ref{case_7a}. 
While both measures penalize the error in \ref{case_7b} more than in \ref{case_7a}, the extra penalization of MGIA is roughly proportional to the distance between $T_1$ and $P_1$, while that 
of the set based measures is not. All set based measures would give the same result even if $P_1$ was a child of $C$, which is a less severe error than when it is a child of $D$. 
Therefore, counting more than once the common paths, may be an advantage of the pair-based measures in some cases. 

\subsection{Very distant predictions}

\begin{figure}[h]
\begin{center}
\subfigure[] {\label{case_8a}
\begin{tikzpicture}[-,>=stealth',shorten >=1pt,auto,node distance=3cm,
  thick,main node/.style={circle,fill=blue!20,draw,font=\sffamily\Large\bfseries, minimum size=11mm, scale=0.45}]

\node[main node] (1) {A};
\node[main node] (2) [below left of=1] {B};
\node[main node] (3) [below right of=1] {C};
\node[main node] (4) [below left of=2] {$T_1$};
\node[main node] (5) [below right of=3] {D};
\node[main node] (6) [below right of=5] {E};
\node[main node] (7) [below right of=6] {$P_2$};
\node[main node] (8) [below left of=6] {$P_1$};

\path[every node/.style={font=\sffamily\small}]
    (1) edge node [left] {} (2)
	edge node [right] {} (3)
    (2) edge node [left] {} (4)		
    (3) edge node [left] {} (5)		
    (5) edge node [right] {} (6)		
    (6) edge node [left] {} (8)
	edge node [right] {} (7);
\end{tikzpicture}
}
\subfigure[] {\label{case_8b}
\begin{tikzpicture}[-,>=stealth',shorten >=1pt,auto,node distance=3cm,
  thick,main node/.style={circle,fill=blue!20,draw,font=\sffamily\Large\bfseries, minimum size=11mm, scale=0.50}]

\node[main node] (1) {0};
\node[main node] (2) [below left of=1] {B};
\node[main node] (3) [below right of=1] {E};
\node[main node] (4) [below left of=2] {$T_1$};
\node[main node] (5) [below left of=3] {$P_1$};
\node[main node] (6) [below right of=3] {$P_2$};

\path[every node/.style={font=\sffamily\small}]
    (1) edge node [left] {} (2)
	edge node [right] {} (3)
    (2) edge node [left] {} (4)		
    (3) edge node [left] {} (5)		
	edge node [right] {} (6);		
    
\end{tikzpicture}
}
\caption{Distant prediction problem}
\label{case8}
\end{center}
\end{figure}
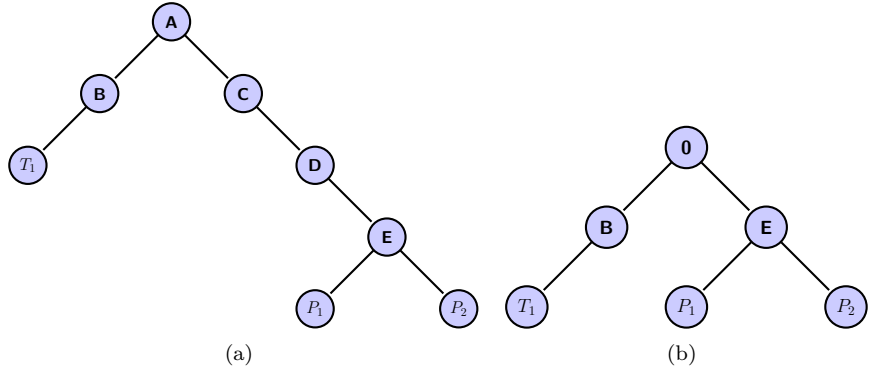

The aim of this case (Figure \ref{case8}) is to show how each of the two types of measure handle very large distances between predicted and true labels. Pair-based measures compute 
the distance between each pair of predicted and true nodes and if this distance is above a certain threshold, a standard maximum distance is assigned. Set-based measures 
can use a threshold on the number of ancestors of the predicted and true nodes that will be used in the augmented sets. Using this threshold, we impose a common ancestor to 
be used in order to connect at least one predicted with one true node, at a distance equal to the threshold.

For example, in the case shown in Figure \ref{case_8a}, if a maximum distance of 4 is used for pair-based measures, then for set-based measures it would be reasonable to request 
a lowest common ancestor at distance 2. By adding the distance of the lowest common ancestor to both the true and the predicted labels, a distance of 4 between them is reached.  

This operation on the hierarchy of Figure \ref{case_8a} leads to the hierarchy of Figure \ref{case_8b}, where an artificial node $0$ was used in order to directly 
connect nodes $B$ and $E$. 
The results of each measure are presented in Table \ref{case8t}. In this example we see that all the measures can be run with a maximum distance 
threshold that might be necessary for computational reasons or due to these long distance problem discussed in section 2.2. Pair-based measures are affected more by 
the threshold than set-based ones, as shown in the example. $l_{\Delta}(Y_{aug},\hat{Y}_{aug})$ decreased by 2 points, while MGIA decreased by 4. 
This is due to multiple counting of paths, which most of the times is undesirable as discussed in section 3.2. 

\begin{table} [h]
\begin{center}
\fitMe{
\begin{tabular}{|c| c c c c c c c c c|}
\hline
& GIE & MGIA & $P_H$ & $R_H$ & $F_H$ & $l_{\Delta}(Y_{aug},\hat{Y}_{aug})$ & $P_{LCA}$ & $R_{LCA}$ & $F_{LCA}$\\
\hline
a & 6 + Max & 0.2(12) & 0.16 & 0.33 & 0.22 & 7 & 0.16 & 0.33 & 0.22\\
b & 4 + Max & 0.466(8) & 0.25 & 0.33 & 0.28 & 5 & 0.25 & 0.33 & 0.28\\
\hline
\end{tabular}
}
\end{center}
\caption{Results per measure for Figure \ref{case8}}
\label{case8t}
\end{table}

\subsection{Over and under-specialization}

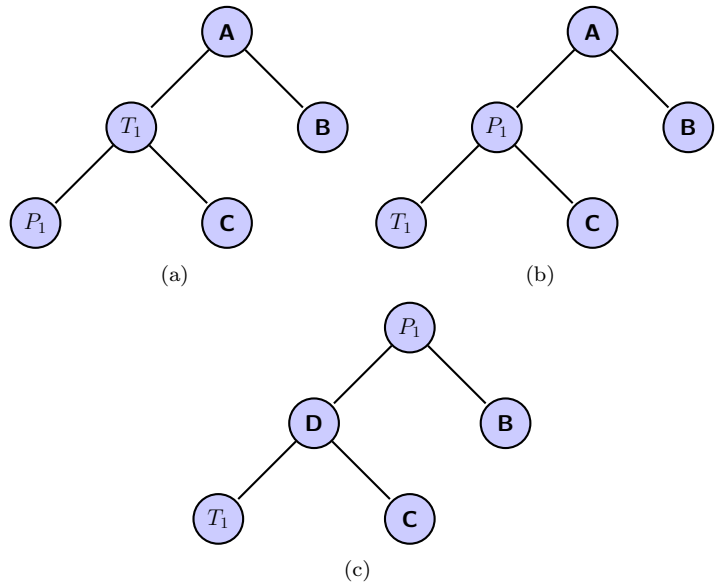
\begin{figure}[h]
\begin{center}
\subfigure[] {\label{case_9a}
  \begin{tikzpicture}[-,>=stealth',shorten >=1pt,auto,node distance=3cm,
    thick,main node/.style={circle,fill=blue!20,draw,font=\sffamily\Large\bfseries, minimum size=11mm, scale=0.6}]

  \node[main node] (1) {A};
  \node[main node] (2) [below left of=1] {$T_1$};
  \node[main node] (3) [below right of=1] {B};
  \node[main node] (4) [below left of=2] {$P_1$};
  \node[main node] (5) [below right of=2] {C};

  \path[every node/.style={font=\sffamily\small}]
      (1) edge node [left] {} (2)
	  edge node [right] {} (3)
      (2) edge node [left] {} (4)
	  edge node [right] {} (5);   	
  \end{tikzpicture}
}
\subfigure[] {\label{case_9b}
  \begin{tikzpicture}[-,>=stealth',shorten >=1pt,auto,node distance=3cm,
    thick,main node/.style={circle,fill=blue!20,draw,font=\sffamily\Large\bfseries, minimum size=11mm, scale=0.6}]

  \node[main node] (1) {A};
  \node[main node] (2) [below left of=1] {$P_1$};
  \node[main node] (3) [below right of=1] {B};
  \node[main node] (4) [below left of=2] {$T_1$};
  \node[main node] (5) [below right of=2] {C};

  \path[every node/.style={font=\sffamily\small}]
      (1) edge node [left] {} (2)
	  edge node [right] {} (3)
      (2) edge node [left] {} (4)
	  edge node [right] {} (5);   	
  \end{tikzpicture}
}
\subfigure[] {\label{case_9c}
  \begin{tikzpicture}[-,>=stealth',shorten >=1pt,auto,node distance=3cm,
    thick,main node/.style={circle,fill=blue!20,draw,font=\sffamily\Large\bfseries, minimum size=11mm, scale=0.6}]

  \node[main node] (1) {$P_1$};
  \node[main node] (2) [below left of=1] {D};
  \node[main node] (3) [below right of=1] {B};
  \node[main node] (4) [below left of=2] {$T_1$};
  \node[main node] (5) [below right of=2] {C};

  \path[every node/.style={font=\sffamily\small}]
      (1) edge node [left] {} (2)
	  edge node [right] {} (3)
      (2) edge node [left] {} (4)
	  edge node [right] {} (5);   	
  \end{tikzpicture}
}
\caption{Simplest over and under-specialization cases}
\label {case9}
\end{center}
\end{figure}

In all the previous cases the predicted and the true categories were leaves of the 
hierarchies, but this is not always the case. Figure \ref{case9} presents simple examples
of an inner node being either a true (Figure \ref{case_9a}) or a predicted 
(Figure \ref{case_9b}) category. As shown in Table \ref{case9t} the two cases receive the same scores.

\begin{table} [h]
\begin{center}
\fitMe{
\begin{tabular}{|c| c c c c c c c c c|}
\hline
& GIE & MGIA & $P_H$ & $R_H$ & $F_H$ & $l_{\Delta}(Y_{aug},\hat{Y}_{aug})$ & $P_{LCA}$ & $R_{LCA}$ & $F_{LCA}$\\
\hline
a & 1 & 0.9(1) & 0.66 & 1 & 0.8 & 1 & 0.5 & 1 & 0.66\\
b & 1 & 0.9(1) & 1 & 0.66 & 0.8 & 1 & 1 & 0.5 & 0.66\\
c & 2 & 0.8(2) & 1 & 0.33 & 0.49 & 2 & 1 & 0.33 & 0.49\\
\hline
\end{tabular}
}
\end{center}
\caption{Results per measure for Figure \ref{case9}}
\label{case9t}
\end{table}

Figure \ref{case_9a} shows a case of over-specialization. As described in 
section 2.2, different evaluation measures treat this type of error  differently.
One could even argue that since $P_1$ is predicted, $T_1$ is also predicted as a direct ancestor of it.
This is not the case here and as shown in Table \ref{case9t} all measures treat it as a misclassification error. 

Regarding under-specialization, the simplest example is shown in Figure \ref{case_9b}. It is 
also considered an error that is more severe the further the true category is from the predicted. 
For example in Figure \ref{case_9c} the predicted node is an ancestor of the predicted node of Figure \ref{case_9b}. 
All measures lead to a higher error estimate in this case than in \ref{case_9b}. A similar example for over-specialization
would lead to the same observations.

\subsection{Summary}

Table \ref{caseAll} presents the advantages and disadvantages of each measure from the scope of the cases presented in Section 2.2. The pair-based measures can handle alternative paths, while, 
from the set-based measures, only the proposed LCA measures are able to deal with them efficiently. All measures can handle over-specialization and under-specialization in some way. All set-based measures
can deal with the pairing problem since they produce augmented sets, while GIE cannot handle it efficiently and this is why MGIA was proposed. Considering the \emph{long distance problem} only pair-based 
measures can handle it by definition, while set-based measures cannot handle it without using the threshold modification that we proposes in Section 3.5. Finally, multi-path counting is a special feature 
of the pair-based measures which although most of the times is undesirable, it could make them behave better than the set-based measures in certain cases. 

As a general conclusion the proposed measures always behave better or at least as well as the existing measures of their category. Therefore, if one wishes to use a pair-based 
or a set-based measure we suggest using the ones proposed in this paper, instead of the existing ones. Furthermore, in most cases one should choose the LCA measures over MGIA, 
due to the multiple counting of paths discussed in section 3.5. Multiple counting of paths is most of the times undesirable since it leads to over-penalization. Additionally, 
these cases could also serve as benchmarks, in order to observe the behaviour of newly proposed hierarchical evaluation measures. In 
this way, we conclude the discussion regarding the behavior of the measures in benchmark cases in order to observe them using real data and systems in the following section.

\begin{table} [h]
\begin{center}
\fitMe{
\begin{tabular}{|c| c c c c c|}
\hline
& GIE & MGIA & $F_H$ & $l_{\Delta}(Y_{aug},\hat{Y}_{aug})$ & $F_{LCA}$\\
\hline
Alternative Paths & + & + & - &  - & +\\
Over-specialization & + & +  & + & + & +\\
Under-specialization & + & + & + & + & +\\
Pairing Problem & - & + &  + & + & +\\
Long Distance Problem & + & + & * & * & *\\
Multiple Path Count & - & - & + & + & +\\
\hline
\end{tabular}
}
\end{center}
\caption{Summary table regarding evaluation measures over certain situations. * means that they cannot handle it their raw versions, but can be modified to do so.}
\label{caseAll}
\end{table}

\section{Empirical Study}

In this section we apply various evaluation measures to the predictions of the systems that participated in the Large Scale Hierarchical Text Classification Pascal Challenges 
of 2011 (LSHTC2) and 2012 (LSHTC3). The goal of this section is to study using real data and systems, the extent to which the performance ranking of systems is affected 
by the choice between flat and hierarchical evaluation 
measures and also by the type of hierarchical measure used. In the first subsection we present the datasets that we used, in the second subsection we discuss the evaluation 
measures included in the comparison and in the final subsection we discuss the results of the study.

In section 3 we demonstrated that, in certain cases, some measures behave more desirably than others. In this section we show that the differences among the methods 
also affect the rankings of real systems in practice.

\subsection{Datasets}

In LSHTC2 three different datasets were provided as three separate tasks. Each participant could participate in any or all of them with the same or with a different system. The first dataset 
(DMOZ) was based on pages crawled from the Open Directory Project (ODP), a human-edited hierarchical directory of the Web.\footnote{http://www.dmoz.org/} The hierarchy of this dataset was transformed into a tree 
and all instances deeper than level five of the hierarchy were transferred to the fifth level, thus leading to a hierarchy with a maximum depth of 5. 
This dataset was the smallest of the three, regarding the number of categories and instances.

The other two datasets of LSHTC2, also used in LSHTC3, are based on DBpedia.\footnote{http://dbpedia.org/About} They are called DBpedia Large and DBPedia Small, respectively.
The largest of the two datasets, DBPedia Large, contains almost all abstracts of the DBpedia, as instances to be used for training and classification, with the exception of some non-English 
abstracts. Therefore this dataset comprises many more categories than the DMOZ one and goes to a larger depth. DBpedia Small is a subset of DBpedia Large, selected in a way that led 
to a dataset of similar size to the DMOZ, while maximizing the ratio of instances per node. This process has resulted in a much easier classification task.The hierarchy of the DBpedia Small 
dataset has been transformed into a $DAG$, by removing cycles, while cycles still appear in DBpedia Large.  

All three datasets were pre-processed in the same way. All the words of the abstracts were stemmed and each stem was mapped to a feature id. The categories (classes) were also mapped to category ids.
Each instance was represented in sparse vector format as a collection of category ids and a collection of feature ids accompanied by their frequencies in the instance. 
The mapping between ids, categories and stems was different for each dataset. Only leaves of each hierarchy were used as valid classification nodes for LSHTC2, 
while in LSHTC3 participants were also allowed to classify instances in inner nodes. For each inner node of the hierarchy that was assigned instances however, a dummy leaf was created for evaluation
purposes as a direct child and all the instances were transferred to the child.

Table \ref{datasets} presents basic statistics of the three datasets. The first two datasets are almost of the same size, but DBpedia Small is more multi-labeled and has a deeper, 
less ballanced hierarchy than DMOZ. However the ratio of training instances to categories is comparable in the two datasets (14.16 for DMOZ and 12.5 for DBpedia Small). 
DBpedia Large is very different in this respect, having a ratio of training instances to categories equal to 7.2. 
Accounting for multi-labelling, this ratio becomes similar in the two DBpedia datasets 
(23.27 for Small and 23.73 for Large) and much smaller for DMOZ (14.5). DBpedia Large is also much larger than the other two datasets in terms of training and testing instances.

\begin{table}[h]
\begin{center}
\fitMe{
\begin{tabular}{|c|c c c|}
\hline
  & DMOZ & DBpedia Small & DBpedia Large\\
\hline
\#cats & 27,875 & 36,504 & 325,056\\  
\#train inst & 394,756 & 456,886 & 2,365,436\\
\#test inst & 104,263 & 81,262 & 452,167\\ 
multi-label factor & 1.0239 & 1.8596 & 3.2614\\
train inst per cat & 14.16 & 12.5 & 7.2\\
multi-label train inst per cat & 14.5 & 23.27 & 23.73\\
max depth & 5 & 10 & 14\\
\hline
\end{tabular}
}
\caption{Basic statistics of datasets showing the number of categories, the number of training and testing instances, the average number of true categories 
per instance (multi-labelling), the ratio of training instances to categories, the ratio of training instances to categories, given multi-labelling and the maximum depth of the graph.}
\label{datasets}
\end{center}
\end{table}

\subsection{Evaluation Measures and Statistical tests}

The evaluation measures that were used in this study were the ones presented in Section 3. Accuracy and GIE are reproduced here as reported during the challenge.
Using these evaluation measures, different rankings of the participating systems are created. In order to measure the correlation 
between these rankings, we used Kendall's rank correlation (\cite{kendall1938}). 

In the LSHTC2 challenge, statistical significance tests were only used for the flat evaluation measures. To the best of our knowledge, the literature does not provide special 
statistical significance tests for hierarchical measures. In this paper, as well as in LSHTC3,  we performed a micro sign test (s-test) similar to that used in (\cite{yang1999re}).
Each of the hierarchical measures provides a score for each instance and this score is always averaged over the number of instances. Assuming that:
\begin{itemize}
\item $a_i$ is the performance of system $a$ for instance $i$, according to an evaluation measure,
\item $b_i$ is the performance of system $b$ for instance $i$, according to the same evaluation measure,
\item $n$ is the number of times that $a_i$ and $b_i$ differ over all $i$,
\item $k$ is the number of times that $a_i$ performs better than $b_i$ over all $i$,
\end{itemize}

the null hypothesis ($H_0$) is that $k$ has a bionomial distribution Bin($n,p$), where $p = 0.5$. $H_1$ is that $p > 0.5$, meaning that system $a$ is better than system $b$.
According to \cite{yang1999re}, if $n$ is greater than 12, which is always the case in these large scale problems, then the p-value can be approximately computed using the standard normal distribution for:
\begin {equation}
Z=\frac{k-0.5n}{0.5\sqrt{n}}
\end {equation}

It is worth stressing that the s-test only takes into account which system performs better at each instance, ignoring how much better it performs. Alternatively, the Wilcoxon signed-rank test \cite{wilcoxon1945individual}
could take into account the difference in performance at each instance. For reasons of simplicity however, in this paper we used the s-test.

\subsection{Results}

In this subsection we present the results for each dataset and discuss the behavior of each measure. Table \ref{dmozresults} presents the results on the DMOZ dataset for all the systems that 
participated in the LSHTC2 challenge. 
Recall that DMOZ has a tree hierarchy and is the least multi-labeled dataset of the three. Systems are evaluated by each measure and are ranked 
in descending order. The number in brackets indicates the system's rank using the corresponding measure. If two systems have the same rank, it means that there is no statistically significant difference
between their results according to our statistical significance tests. Table \ref{dmozkendal} presents the Kendall rank correlation 
between each pair of rankings.

The first observation is that the ranking of the flat accuracy is different from that of the other hierarchical measures. This shows that flat and hierarchical measures 
treat the problem differently. Another interesting observation is that the rankings also differ between hierarchical measures.

The handling of multiple labels per instance is an important aspect the classification methods. Table \ref{dmozStats} presents the average number of predictions per instance for each system. 
Since most instances of the dataset are single-labeled, most of the participants treated the task as a single-label one. As discussed in previous sections the treatment of
multi-labeling by different measures, greatly affects their behavior, but since multi-labeling is rare in this dataset, this decision did not affect much the hierarchical measures. 
However there are some examples of systems, such as M and J, which assign multiple labels and perform better according to hierarchical measures than according to accuracy. D and E, on the other hand, perform 
worse using some hierarchical measures than with accuracy. The more multi-labeled a result is, the greater the opportunity is for a hierarchical measure to reward or penalize the
systems for its decision.

As discussed in previous sections, hierarchical measures vary in the way they handle multi-labeling and $DAG$ hierarchies. Being a tree hierarchy and almost single-labeled, 
the DMOZ dataset does not reveal a lot of these differences. The tree hierarchy is the main reason why, according to Table \ref{dmozkendal}, $l_{\Delta}(Y_{aug},\hat{Y}_{aug})$ and $F_H$ are very highly 
correlated with $F_{LCA}$, since their main difference is in the way they treat multiple ancestors, something possible only in DAGs and not in trees. $F_{LCA}$ and $F_H$ are more correlated with each other than with 
$l_{\Delta}(Y_{aug},\hat{Y}_{aug})$, as expected, since they differ in the calculations they perform on the augmented sets. We also observe a correlation between GIE and MGIA, although the main reasons here are not the 
hierarchy, but the limited multi-labeling that provides fewer opportunities for dealing with the pairing problem and the proposed tranformation of the error that our MGIA performs.

\begin{table}[h]
\begin{center}
\fitMe{
\begin{tabular}{|c|c c c c c c|}
\hline
System & Acc  & GIE & $F_H$ & $l_{\Delta}(Y_{aug},\hat{Y}_{aug})$ & MGIA & $F_{LCA}$\\
\hline
A & 0.388 (1) & 2.829 (2) & 0.653 (3) & 3.962 (2) & 0.541 (2) & 0.557 (2) \\
B & 0.387 (2) & 2.823 (1) & 0.660 (1) & 3.910 (1) & 0.550 (1) & 0.559 (1) \\
C & 0.386 (3) & 2.831 (3) & 0.654 (2) & 3.987 (3) & 0.542 (1) & 0.555 (3) \\
D & 0.380 (4) & 3.322 (6) & 0.642 (5) & 4.257 (4) & 0.515 (4) & 0.544 (5) \\
E & 0.378 (5) & \textbf{3.832 (10)} & 0.640 (6) & 4.458 (7) & 0.501 (5) & 0.538 (6)\\
F & 0.371 (6) & 2.891 (4) & 0.652 (4) & 3.996 (4) & \textbf{0.538 (3)} & \textbf{0.547 (4)}\\
G & 0.347 (7) & 3.027 (5) & 0.622 (7) & 4.335 (6) & 0.497 (6) & 0.522 (7)\\
H & 0.284 (8) & 3.456 (7) & \textbf{0.497 (11)} & \textbf{5.878 (11)} & 0.364 (8) & 0.440 (9)\\
I & 0.269 (9) & 3.503 (9) & 0.571 (8) & 4.987 (9) & 0.421 (7) & 0.460 (8)\\
J & 0.262 (10) & 3.476 (8) & 0.570 (8) & 4.966 (8) & 0.428 (7) & 0.458 (8)\\
K & 0.172 (11) & 3.898 (11) & 0.469 (12) & 6.165 (12) & 0.318 (10) & 0.373 (11)\\
L & 0.155 (12) & 4.010 (12) & 0.446 (13) & 6.430 (13) & 0.282 (11) & 0.353 (12)\\
M & 0.153 (13) & 4.024 (13) & 0.497 (10) & 5.803 (10) & 0.333 (9) & 0.374 (10)\\
N & 0.107 (14) & 4.289 (14) & 0.384 (14) & 7.080 (14) & 0.202 (12) & 0.306 (13)\\
O & 0.087 (15) & 4.419 (15) & 0.340 (15) & 7.744 (15) & 0.175 (13) & 0.280 (14)\\
\hline 
\end{tabular}
}
\caption{DMOZ results in LSHTC2. Interesting rank changes are marked with bold.}
\label{dmozresults}
\end{center}
\end{table}

\begin{table}[h]
\begin{center}

\begin{tabular}{|c| c c c c c c|}
\hline
& Acc  & GIE & $F_H$ & $l_{\Delta}(Y_{aug},\hat{Y}_{aug})$ & MGIA & $F_{LCA}$\\
\hline
Acc & 1 &  &  &  &  & \\
GIE & 0.829 & 1 &  &  &  & \\
$F_H$ & 0.842 & 0.785 & 1 &  &  & \\
$l_{\Delta}(Y_{aug},\hat{Y}_{aug})$ & 0.790 & 0.810 & 0.919 & 1 &  & \\
MGIA & 0.829 & 0.810 & 0.976 & 0.924 & 1 & \\
$F_{LCA}$ & 0.867 & 0.810 & 0.976 & 0.924 & 0.962 & 1\\
\hline 
\end{tabular}

\caption{Kendall's rank correlation on the evaluation measure rankings of DMOZ in LSHTC2.}
\label{dmozkendal}
\end{center}
\end{table}

\begin{table}[h]
\begin{center}
\fitMe{
\begin{tabular}{|c c c c c c c c c c c c c c c|}
\hline
A & B & C & D & E & F & G & H & I & J & K & L & M & N & O\\
1 & 1 & 1 & 1.11 & 1.22 & 1 & 1 & 1 & 1.02 & 1.01 & 1 & 1 & 1.02 & 1 & 1 \\
\hline 
\end{tabular}}
\caption{Average number of predictions per instance of DMOZ systems in LSHTC2.}
\label{dmozStats}
\end{center}
\end{table}

Tables \ref{wikipediasmallresults} and \ref{wikismallkendal} present the LSHTC2 results on DBpedia Small, which is a more multi-labeled dataset with a $DAG$ hierarchy. 
As expected, these characteristics greatly affect the behavior of the measures. The most important observation is that the two hierarchical measures (GIE and $l_{\Delta}(Y_{aug},\hat{Y}_{aug})$) 
that measure only the error, without performing a transformation to it, have very low correlation with Acc and the other hierarchical measures. Furthermore, taking into consideration the average 
predictions per instance of each system (Table \ref{wikipediasmallStats}), we observe a relation between the rankings of these two measures. By computing only the error these two measures take 
into account only FP and FN, without counting the TP. For this reason they tend to penalize systems with higher average predictions per instance, since they are more likely to make more mistakes. 
The way the rest of the set based measures handle the augmented true and predicted sets of classes 
and the transformation that MGIA performs to the error is much closer to the idea of doing calculations with TP, FP, TN, FN, as accuracy does and for this reason these measures are more correlated with it. These measures also penalize less the systems that have a higher average predictions per instance, if they manage to have some extra TPs by doing so.

\begin{table}[h]
\begin{center}
\fitMe{
\begin{tabular}{|c|c c c c c c|}
\hline
System & Acc  & GIE & $F_H$ & $l_{\Delta}(Y_{aug},\hat{Y}_{aug})$ & MGIA & $F_{LCA}$\\
\hline
A2 & 0.374 (1) & \textbf{4.171 (5)} & 0.647 (1) & \textbf{12.114 (6)} & 0.356 (1)& 0.481 (1)\\
B2 & 0.362 (2) & \textbf{4.364 (6)} & 0.641 (3) & \textbf{12.000 (4)} & 0.337 (2) & 0.470 (2) \\
C2 & 0.354 (3) & 4.076 (4) & 0.646 (2) & 11.651(3) & 0.323 (4) & 0.463 (3)\\
D2 & 0.351 (4) & 3.858 (2) & 0.629 (4) & 11.332 (2) & 0.329 (3) & 0.462 (4) \\
E2 & 0.279 (5) & 3.726 (1) & 0.600 (5) & 11.326 (1) & 0.286 (5) & 0.414 (5)\\
F2 & 0.252 (6) & 3.859 (3) & 0.579 (6) & 11.996 (5) & 0.280 (6) & 0.399 (6) \\
G2 & 0.249 (7) & 5.701 (7) & 0.561 (7) & 16.915 (7) & 0.245 (7) & 0.381 (7)\\
\hline 
\end{tabular}
}
\caption{DBpedia Small results in LSHTC2. Significant deviations of rankings are highlighted in bold.}
\label{wikipediasmallresults}
\end{center}
\end{table}

\begin{table}[h]
\begin{center}

\begin{tabular}{|c| c c c c c c|}
\hline
& Acc  & GIE & $F_H$ & $l_{\Delta}(Y_{aug},\hat{Y}_{aug})$ & MGIA & $F_{LCA}$\\
\hline
Acc & 1 &  &  &  & & \\
GIE & -0.143 & 1 &  &  &  & \\
$F_H$ & 0.905 & -0.048 & 1 &  & & \\
$l_{\Delta}(Y_{aug},\hat{Y}_{aug})$ & -0.143 & 0.810 & -0.048 & 1 &  & \\
MGIA & 0.810 & -0.143 & 0.714 & -0.143 & 1 & \\
$F_{LCA}$ & 0.905 & -0.238 & 0.810 & -0.238 & 0.905 & 1\\
\hline 
\end{tabular}

\caption{Kendall's rank correlation on the evaluation measure rankings of DBpedia Small in LSHTC2.}
\label{wikismallkendal}
\end{center}
\end{table}

\begin{table}[h]
\begin{center}
\begin{tabular}{|c c c c c c c|}
\hline
A2 & B2 & C2 & D2 & E2 & F2 & G2\\
2.04 & 1.94 & 1.82 & 1.51 & 1.11 & 1.14 & 2.84\\
\hline 
\end{tabular}
\caption{Average predictions per instance of DBpedia Small systems in LSHTC2.}
\label{wikipediasmallStats}
\end{center}
\end{table}

Tables \ref{wikipediasmallresults3} and \ref{wikismallkendal3} present the results on the same dataset (DBpedia Small), but with the systems of LSHTC3. 
The number of systems participating in LSHTC3 is much larger than LSHTC2 (17 instead of 7). 
This is not only important for statistical reasons (more experiments lead to safer conclusions), but also because according to Table \ref{wikipediasmallStats3} 
we now have more systems with higher average number of predictions per instance, something which affects the behavior of the measures. Another important difference is 
that in LSHTC3 systems were allowed to classify to inner nodes, even if these nodes did not have any training instance directly belonging to them. 

$F_H$ and $F_{LCA}$ are the hierarchical measure that are most correlated with 
flat accuracy, although the correlation is much lower in this case where we have many more systems and inner node classification is treated as a mistake by accuracy. 
The correlation between GIE and MGIA is much higher than that of LSHTC2, but they are not fully correlated. 
A high correlation also continues to be observed between GIE and $l_{\Delta}(Y_{aug},\hat{Y}_{aug})$ for the reason explained previously in LSHTC2.

A very interesting case is that of system $X2$, which predicts many categories (labels) per instance, 10.649 labels per instance, when the average 
true labels per instance is 1.8550. This means that a large number of predicted labels is wrong, while the predicted labels could still be in the vicinity of the correct ones. 
As expected, flat accuracy penalizes this behavior giving the lowest rank to this system. GIE and MGIA also penalize this system, although MGIA less severely. On the other hand, the set-based measures 
do not punish X2 that much (6 for $F_H$, 4 for $l_{\Delta}(Y_{aug},\hat{Y}_{aug})$ and 9 for $F_{LCA}$.
A closer look at the system shows that it is in fact not that bad. However, it returns all the nodes of a path from the root to leaf as predicted labels, instead of just the leaf. 
Set-based measures still penalize it when the leaf and its ancestors are wrong predictions, but they do not over-penalize it, unlike pair-based measures. This is a nice example, 
of the difference between set-based and pair-based measures. We can also observe that MGIA and $F_{LCA}$ are the less extreme measures of their categories to a point where the ranks are very close
10 and 9 respectively. This is because these measures can be seen as hybrid measures since the first also conducts a set operation (although it is a pair-based measure) and the second does some kind
of matching between true and predicted nodes in order to create the augmented sets. These characteristics help them overcome the weaknesses of the measures of their respective categories and in that way their behavior is more desirable.

\begin{table}[h]
\begin{center}
\fitMe{
\begin{tabular}{|c|c c c c c c|}
\hline
System & Acc  & GIE & $F_H$ & $l_{\Delta}(Y_{aug},\hat{Y}_{aug})$ & MGIA & $F_{LCA}$\\
\hline
H2 & 0.438 (1) & 3.060 (1) & 0.709 (1) & 9.096 (1) & 0.421 (2) & 0.543 (1)\\
I2 & 0.429 (2) & 3.155 (2) & 0.689 (3) & 9.310 (2) & \textbf{0.398 (4)} & \textbf{0.525	(4)}\\
J2 & 0.42 (3) &	3.530 (5) & 0.692 (2) & 10.143 (6) & 0.403 (3) & 0.529 (2)\\
K2 & 0.417 (4) & \textbf{4.428	(11)} & 0.677 (5) & 11.385 (11) & 0.378 (6) & 0.509 (5)\\
L2 & 0.408 (5) & 3.187	(3) & 0.680 (4) & 9.561	(3) & \textbf{0.443 (1)} & \textbf{0.527 (3)}\\
M2 & 0.385 (6) & 3.319	(4) & 0.666 (7)	& 10.122 (5) & 0.390 (4) & 0.500 (6)\\
N2 & 0.371 (7) & 4.991	(13) & 0.645 (8) & 13.117 (13) & 0.342 (8) & 0.476 (8)\\
O2 & 0.357 (8) & 4.302	(10) & 0.643 (8) & 12.185 (12) & 0.323	(9) & 0.462 (10)\\
P2 & 0.354 (9) & 3.550	(6) & 0.633 (9) & \textbf{11.146 (8)} & 0.381 (5) & 0.478 (7)\\
Q2 & 0.327 (10) & 3.600	(8) & 0.639 (8)	& 10.944 (7) & 0.312 (11) & 0.450 (12)\\
R2 & 0.32 (11) & 3.552	(7) & 0.603 (10) & 11.365 (10) & \textbf{0.361 (7)} & 0.453 (11)\\
S2 & 0.298 (12) & 5.693	(14) & 0.549 (13) & 16.873 (15) & 0.243 (13) & 0.407 (13)\\
T2 & 0.25 (13) & 3.741	(9) & 0.592 (11)& 11.304 (9) & \textbf{0.089 (14)} & 0.397 (14)\\
U2 & 0.249 (14) & 5.701	(15) & 0.561 (12) & 16.915 (16)	& 0.245 (12) & 0.381 (15)\\
V2 & 0.245 (15) & 4.780	(12) & 0.537 (14) & 14.351 (14) & 0.234	(13) & 0.374 (16)\\
W2 & 0.063 (16) & 9.139	(16) & 0.345 (15) & 24.009 (17) & 0.045 (15) &	0.208 (17)\\
X2 & \textbf{0.047 (17)} & \textbf{25.775 (17)} & \textbf{0.668 (6)} & \textbf{9.607 (4)} & \textbf{0.321 (10)} & \textbf{0.471 (9)}\\
\hline 
\end{tabular}
}
\caption{DBpedia Small results in LSHTC3. With bold, interesting differences in rankings.}
\label{wikipediasmallresults3}
\end{center}
\end{table}

\begin{table}[h]
\begin{center}

\begin{tabular}{|c| c c c c c c|}
\hline
& Acc  & GIE & $F_H$ & $l_{\Delta}(Y_{aug},\hat{Y}_{aug})$ & MGIA & $F_{LCA}$\\
\hline
Acc & 1 &  &  &  & & \\
GIE & 0.662 & 1 &  &  &  & \\
$F_H$ & 0.765 & 0.485 & 1 &  & & \\
$l_{\Delta}(Y_{aug},\hat{Y}_{aug})$ & 0.485 &0.735 & 0.662 & 1 &  & \\
MGIA & 0.691 & 0.618 & 0.721 & 0.588 & 1 & \\
$F_{LCA}$ & 0.794 & 0.574 & 0.853 & 0.603 & 0.838 & 1\\
\hline 
\end{tabular}

\caption{Kendall's rank correlation on the evaluation measure rankings of DBpedia Small in LSHTC3.}
\label{wikismallkendal3}
\end{center}
\end{table}

\begin{table}[h]
\begin{center}

\begin{tabular}{|c c c c c c c c c|}
\hline
H2 & I2 & J2 & K2 & L2 & M2 & N2 & O2 & P2\\
1.506 & 1.482 & 1.909 & 2.208 & 1.415 & 1.490 & 2.427 & 1.889 & 1.414\\
\hline
Q2 & R2 & S2 & T2 & U2 & V2 & W2 & X2 & \\
1.529 & 1.184 & 2.423 & 2.000 & 2.841 & 1.712 & 4.334 & 10.649 & \\
\hline 
\end{tabular}

\caption{Average predictions per instance on DBpedia Small in LSHTC3.}
\label{wikipediasmallStats3}
\end{center}
\end{table}

On the third dataset (DBpedia Large) we faced some computational issues with the hierarchical measures. The problem originated from the very large scale of the dataset's hierarchy, which is a 
$DAG$ (in reality it contains circles but we remove them). To avoid the computational problems, we run the evaluation measures with a maximum path threshold of 2 and 4. This means that all 
nodes are forced to have a lowest common ancestor at a depth of 1 and 2 respectively (if they do not have one we create a dummy one). Although this seems restrictive, 
it is very similar to the idea behind the Long Distance problem of Figure \ref{fig:ph5} discussed in section 2. In the Long Distance problem we used dummy nodes in order to 
link nodes that were further than a threshold from each other, in order to avoid overpenalization. The same dummy nodes are used here for computational reasons.

Tables \ref{wikipedialargeresults4} and \ref{wikilargekendal4} present the results for a distance threshold of 4 for the systems of LSHTC2. Since this dataset has the most 
complex hierarchy and it is the most multi-labeled one, it should be treated very differently by each measure. Interestingly the rankings of $F_H$, MGIA and $F_{LCA}$
remain highly correlated with accuracy compared to the other measures, although the number of systems is not high enough (only 5 systems) in order to make safe conclusions.

Another interesting observation is the disagreement of  GIE and MGIA about systems C3 and E3. As shown in Table \ref{wikipedialargeStats}, E3 predicts fewer 
categories per instance than C3. Since most of the times the predicted categories (labels) are fewer than the true ones and GIE over-penalizes all the unmatched true categories, 
it is natural for GIE to penalize system C3 more than E3. This problem is fixed by MGIA, which allows multi-pairing and this is why it instead ranks C3 
as a better system than E3.
We also notice that this difficult hierarchy affects the performance of $l_{\Delta}(Y_{aug},\hat{Y}_{aug})$ and its ranks become less correlated with $F_{LCA}$. 
A more interesting observation is that $F_H$, MGIA and $F_{LCA}$ are completely correlated with each other and not correlated with the average number of predictions per instance 
(Table \ref{wikipedialargeStats}).

\begin{table}[h]
\begin{center}
\fitMe{
\begin{tabular}{|c|c c c c c c|}
\hline
System & Acc  & GIE & $F_H$ & $l_{\Delta}(Y_{aug},\hat{Y}_{aug})$ & MGIA & $F_{LCA}$\\
\hline
A3 & 0.347 (1) & 4.647 (4) & 0.538 (1) & 42.470 (3) & 0.319 (1) & 0.44 (1)\\
B3 & 0.337 (2) & 4.392 (2) & 0.511 (2) & 40.811 (1) & 0.315 (2) & 0.437 (2)\\
C3 & 0.283 (3) & \textbf{6.178 (5)} & 0.440 (4) & 50.709 (5) & \textbf{0.253 (4)} & 0.39 (4)\\
D3 & 0.272 (4) & 4.288 (1) & 0.483 (3) & 42.430 (2) & 0.294 (3) & 0.388 (3)\\
E3 & 0.177 (5) & \textbf{4.535 (3)} & 0.314 (5) & 47.957 (4) & \textbf{0.212 (5)} & 0.331 (5)\\
\hline 
\end{tabular}
}
\caption{DBpedia Large results with a maximum path threshold of 4 in LSHTC2. With bold, interesting differences in rankings.}
\label{wikipedialargeresults4}
\end{center}
\end{table}

\begin{table}[h]
\begin{center}

\begin{tabular}{|c| c c c c c c|}
\hline
& Acc  & GIE & $F_H$ & $l_{\Delta}(Y_{aug},\hat{Y}_{aug})$ & MGIA & $F_{LCA}$\\
\hline
Acc & 1 & & & & &\\
GIE & 0 & 1 & & & & \\
$F_H$ & 0.8 & 0.2 & 1 & &  &\\
$l_{\Delta}(Y_{aug},\hat{Y}_{aug})$ & 0.2 & 0.8 & 0.4 & 1 & & \\
MGIA & 0.8 & 0.2 & 1 & 0.4 & 1 & \\
$F_{LCA}$ & 0.8 & 0.2 & 1 & 0.4 & 1 & 1\\
\hline 
\end{tabular}

\caption{Kendall's rank correlation on the evaluation measure rankings with a maximum path threshold of 4 on DBpedia Large in LSHTC2.}
\label{wikilargekendal4}
\end{center}
\end{table}

\begin{table}[h]
\begin{center}
\begin{tabular}{|c c c c c|}
\hline
 A3 &  B3 & C3 & D3 & E3\\
 3.15 & 2.69 & 3.62 & 2.81 & 1.27\\
\hline 
\end{tabular}
\caption{Average predictions per instance on DBpedia Large systems in LSHTC2.}
\label{wikipedialargeStats}
\end{center}
\end{table}

Tables \ref{wikipedialargeresults2} and \ref{wikilargekendal2} present the results for a maximum path threshold of 2. 
The main purpose of this experiment is to show that the measures remain largely unaffected by this parameter. 
Indeed most rankings do not seem to be affected compared to Tables \ref{wikipedialargeresults4} and \ref{wikilargekendal4}. Nevertheless, general advice is to 
keep the maximum paths parameter as large as possible.

\begin{table}[h]
\begin{center}
\fitMe{
\begin{tabular}{|c|c c c c c c|}
\hline
& Acc  & GIE & $F_H$ & $l_{\Delta}(Y_{aug},\hat{Y}_{aug})$ & MGIA & $F_{LCA}$\\
\hline
A3 & 0.347 (1) & 4.647 (4) & 0.503 (1) & 22.006 (3) & 0.206 (2) & 0.460 (2)\\
B3 & 0.337 (2) & 4.392 (2) & 0.475 (2) & 21.028 (1) & 0.207 (1) & 0.461 (1)\\
C3 & 0.283 (3) & 6.178 (5) & 0.412 (4) & 25.465 (5) & 0.163 (3) & 0.416 (3)\\
D3 & 0.272 (4) & 4.288 (1) & 0.439 (3) & 21.702 (2) & 0.155 (4) & 0.405 (4)\\
E3 & 0.177 (5) & 4.535 (3) & 0.282 (5) & 24.146 (4) & 0.134 (5) & 0.360 (5)\\
\hline 
\end{tabular}
}
\caption{DBpedia Large results with a maximum path threshold of 2 in LSHTC2.}
\label{wikipedialargeresults2}
\end{center}
\end{table}

\begin{table}[h]
\begin{center}

\begin{tabular}{|c| c c c c c c|}
\hline
& Acc  & GIE & $F_H$ & $l_{\Delta}(Y_{aug},\hat{Y}_{aug})$ & MGIA & $F_{LCA}$\\
\hline
Acc & 1 & &  & &  & \\
GIE & 0 & 1 &  &  &  & \\
$F_H$ & 0.8 & 0.2 & 1 & &  & \\
$l_{\Delta}(Y_{aug},\hat{Y}_{aug})$ & 0.2 & 0.8 & 0.4 & 1 & & \\
MGIA & 0.8 & 0.2 & 0.6 & 0.4 & 1 & \\
$F_{LCA}$ & 0.8 & 0.2 & 0.6 & 0.4 & 1 & 1\\
\hline 
\end{tabular}

\caption{Kendall's rank correlation on the evaluation measure rankings with a maximum path threshold of 2 on DBpedia Large in LSHTC2.}
\label{wikilargekendal2}
\end{center}
\end{table}

Tables \ref{wikipedialargeresultsLSHCT3} and \ref{wikilargekendalLSHCT3} present the results on the same dataset (DBpedia Large), but with the systems of LSHTC3. 
The main difference is that in LSHTC3, systems were allowed to classify to inner nodes. Table \ref{wikipedialargeStatsLSHCT3} shows that the average number of predictions 
per instance is similar to that of LSHTC2. The most interesting observation is that while system F3 is ranked once again high according to Accuracy and all the other 
hierarchical measures except GIE and $l_{\Delta}(Y_{aug},\hat{Y}_{aug})$ which rank it as one of the worst systems. 
It is even more interesting that, according to Table \ref{wikipedialargeStatsLSHCT3}, 
system F3 provides the most labels per instance. The assignement of many labels is penalized heavily by measures based only on FP and FN as we mentioned before. 

Another interesting observation is that MGIA and $F_{LCA}$ are not fully correlated anymore. In fact MGIA is more correlated with 
$F_H$ than with $F_{LCA}$. As the hierarchy becomes more complicated and the results more multi-labeled our two proposed measures
behave more differently. Finally system G3, which predicts the smallest number of instances per document, is one ofthe best systems according to all measures except 
MGIA which ranks it as one of worst. This is because although G3 has the highest $P_H$ and $P_{LCA}$, it also has a very low $R_H$ and $R_{LCA}$ compared to other systems.
The computation of $F_1$ seems more sutable in this case compared to the transformation that we proposed for MGIA, one extra reason why we propose $F_{LCA}$ over MGIA.

Similar experiments with a maximum path threshold of 2 were also conducted with systems of LHSTC3, but the results were similar to the ones of LSHTC2 and thus we omit them here, in 
the interest of space.
\begin{table}[h]
\begin{center}
\fitMe{
\begin{tabular}{|c|c c c c c c|}
\hline
System & Acc  & GIE & $F_H$ & $l_{\Delta}(Y_{aug},\hat{Y}_{aug})$ & MGIA & $F_{LCA}$\\
\hline
F3 & \textbf{0.381 (1)} & \textbf{6.104 (6)} & 0.557 (1) & \textbf{42.790 (5)} & 0.374 (2) & 0.465 (1)\\
G3 & 0.346 (2) & 3.756 (1) & 0.513 (3) & 33.630 (1) & \textbf{0.309 (5)} & 0.456 (2)\\
H3 & 0.340 (3) & 3.763 (2) & 0.508 (4) & 38.137 (2) & 0.35 (3) & 0.45 (3)\\
I3 & 0.333 (4) & 3.763 (3) & 0.507 (5) & 44.435 (6) & 0.31 (4) & \textbf{0.43 (5)}\\
J3 & 0.332 (5) & 4.216 (4) & 0.517 (2) & 40.560 (3) & 0.381 (1) & 0.449 (4)\\
K3 & 0.272 (6) & 4.288 (5) & 0.483 (6) & 42.430 (4) & 0.294 (6) & 0.388 (6)\\
\hline 
\end{tabular}
}
\caption{DBpedia Large results with a maximum path threshold of 4 in LSHTC3. With bold, interesting differences in rankings.}
\label{wikipedialargeresultsLSHCT3}
\end{center}
\end{table}

\begin{table}[h]
\begin{center}

\begin{tabular}{|c| c c c c c c|}
\hline
& Acc  & GIE & $F_H$ & $l_{\Delta}(Y_{aug},\hat{Y}_{aug})$ & MGIA & $F_{LCA}$\\
\hline
Acc & 1 & & & & &\\
GIE & 0.276 & 1 & & & & \\
$F_H$ & 0.6 & 0.138 & 1 & & & \\
$l_{\Delta}(Y_{aug},\hat{Y}_{aug})$ & 0.2 & 0.552 & 0.067 & 1 & & \\
MGIA & 0.2 & -0.276 & 0.6 & -0.067 & 1 & \\
$F_{LCA}$ & 0.828 & 0.071 & 0.828 & 0.276 & 0.414 & 1\\
\hline 
\end{tabular}
\caption{Kendall's rank correlation on the evaluation measure rankings with a maximum path threshold of 4 on DBpedia Large in LSHTC3.}
\label{wikilargekendalLSHCT3}
\end{center}
\end{table}

\begin{table}[h]
\begin{center}
\begin{tabular}{|c c c c c c|}
\hline
F3 &  G3 & H3 & I3 & J3 & K3\\
3.949 & 1.482 & 2.315 & 2.903 & 2.902 & 2.810\\
\hline 
\end{tabular}
\caption{Average predictions per instance of DBpedia Large systems in LSHTC3.}
\label{wikipedialargeStatsLSHCT3}
\end{center}
\end{table}

The experiments presented in this section illustrated, with the use of real systems and datasets, that hierarchical measures treat the competing systems differently than flat 
measures. This was shown by presenting the differences in the rankings of the systems across the three datasets. Flat evaluation measures, which are commonly used, often provide
a false indication of which system performs better by ignoring hierarchical dependencies of classes and treating all errors equally.
As a result, their use guides research away from the methods that incorporate the hierarchy in the classification process. 
We also showed that different variants of hierarchical measures give different rankings under different conditions. The goal was not to choose the best measure, 
but to show that different hierarchical evaluation measures give different results, not only in absolute values but 
also in the ranking of the systems. Finally we showed that the scale of the task is also an issue which requires attention.
                                                                                                                                                                                                     
\section{Conclusions}
In this work we studied the problem of evaluating the performance of hierarchical classification methods.
Specifically, this work abstracted and presented the key points of existing performance measures.
We proposed a grouping of the methods into a) \emph{pair-based} and b) \emph{set-based}. Measures in the former 
group attempt to match each prediction to a true class and measure their distance. 
In contrast set-based measures use the hierarchical relations in order to augment the 
sets of predicted and true labels, and then use set operations, like symmetric difference and intersection,
on the augmented label sets.

In order to model pair-based measures, we introduced a novel generic framework based on flow networks, while 
for set-based measures we provided a framework
based on set operations. Thus, salient features of these measures are stressed and 
presented under a common formalism.

Another contribution of this paper was the proposal of two measures (one for each group) that address
several deficiencies of existing measures. The proposed measures, along with existing ones were assessed in two ways.
First, we applied them to selected cases, in order to demonstrate their pros and cons. Second, we studied them empirically on three large
datasets based on DMOZ and Wikipedia with different characteristics (single-label, multi-label,
tree and DAG hierarchies). The analysis of the results showed that the hierarchical measures behave
differently, especially in cases of multi-label data and DAG hierarchies. Also, the two proposed
measures have shown a more robust behavior compared to their counterparts.
Finally, the results supported our initial premise that flat measures are not adequate for
evaluating hierarchical categorization systems.

Our analysis showed that although in certain rare cases pair-based measures may behave more desirably, in most cases the set-based method proposed in
this paper ($F_{LCA}$) exhibits the most desirable behavior than that of our proposed pair-based measure (MGIA), since it is actually a hybrid measure 
because of the pairing way that the LCAs are selected. This why we propose the use of $F_{LCA}$ instead of all other hierarchical measures, although it 
is still an open-issue to propose a measure that combines all the pros of both our proposed MGIA and $F_{LCA}$.

\end{document}